\documentclass[a4paper,10pt]{article}

\usepackage[a4paper,left=3cm,right=3cm,top=2.5cm,bottom=2.5cm]{geometry}
\usepackage{hyperref}
\usepackage{subfig}
\usepackage{pgfplots}
\usepackage{pgfplotstable}
\usepackage{mathtools}
\usepackage{multicol}
\usepackage{comment}
\usepackage{booktabs}
\pgfplotsset{compat=1.5}
\usepackage{amssymb}
\usepackage{url}
\usepackage{bm}

\usepackage{pdfsync}
\usepackage{float}
\usepackage{tabularx}
\usepackage{enumerate}
\usepackage{array}
\usepackage{xspace}
\usepackage{siunitx}
\usepackage{authblk}

\usepackage{amsthm}
\usepackage{amsmath}
\usepackage{stmaryrd}
\usepackage{graphicx}
\usepackage{booktabs}
\usepackage{cleveref}

\usepackage{dsfont}
\usepackage{algorithm}
\usepackage[noend]{algpseudocode}
\usepackage{listings}
\usepackage{subfig}
\usepackage{pgfplots}
\usepackage{tikz}
\usetikzlibrary{shapes,arrows,positioning,chains,fit,arrows.meta,backgrounds}

\tikzstyle{P_node} = [circle,draw=darkgray,thick,align=center,minimum size=0.9cm,fill=cyan!50]
\tikzstyle{O_node} = [circle,draw=darkgray,line width=.6mm,align=center,minimum size=0.9cm,fill=cyan!20]
\tikzstyle{D_node} = [circle,draw=darkgray,thick,align=center,minimum size=0.9cm,fill=green!70!cyan!30]
\tikzstyle{D_NN_node} = [circle,draw=darkgray,thick,align=center,minimum size=0.9cm,fill=green!60!cyan!15]
\tikzstyle{U_node_prob} = [circle, draw=darkgray, thick, minimum size=0.9cm,fill=red!40]
\tikzstyle{U_node_act} = [rectangle, draw=darkgray, line width=.6mm, minimum width=0.9cm, minimum height=0.9cm,fill=red!40]
\tikzstyle{Q_node} = [circle,draw=darkgray,thick,align=center,minimum size=0.9cm,fill=magenta!30]
\tikzstyle{R_node} = [rectangle,draw=darkgray,thick,align=center,minimum width=0.85cm, minimum height=0.85cm,fill=yellow!30,rotate=45]

\DeclareMathOperator*{\argmax}{arg\,max}

\newcommand{\squeezeup}{\vspace{-2.mm}}

\begin{document}

\title{Adaptive digital twins for predictive decision-making:\\Online Bayesian learning of transition dynamics}

\author[1]{Eugenio~Varetti\footnote{eugenio.varetti@mail.polimi.it}}
\author[2]{Matteo~Torzoni\footnote{matteo.torzoni@polimi.it}}
\author[3]{Marco~Tezzele\footnote{marco.tezzele@emory.edu}}
\author[1]{Andrea~Manzoni\footnote{andrea1.manzoni@polimi.it}}

\affil[1]{MOX -- Dipartimento di Matematica, Politecnico di Milano, Milan, 20133, Italy}
\affil[2]{Dipartimento di Ingegneria Civile e Ambientale, Politecnico di Milano, Milan, 20133, Italy}
\affil[3]{Mathematics Department, Emory University, Atlanta, 30322, GA, United States}

\maketitle
\squeezeup
\begin{abstract}
This work shows how adaptivity can enhance value realization of digital twins in civil engineering. We focus on adapting the state transition models within digital twins represented through probabilistic graphical models. The bi-directional interaction between the physical and virtual domains is modeled using dynamic Bayesian networks. By treating state transition probabilities as random variables endowed with conjugate priors, we enable hierarchical online learning of transition dynamics from a state to another through effortless Bayesian updates. We provide the mathematical framework to account for a larger class of distributions with respect to the current literature on digital twins. To compute dynamic policies with precision updates we solve parametric Markov decision processes through reinforcement learning. The proposed adaptive digital twin framework enjoys enhanced personalization, increased robustness, and improved cost-effectiveness. We assess our approach on a case study involving structural health monitoring and maintenance planning of a railway bridge.
\end{abstract}

\tableofcontents

\section{Introduction}
\label{sec:introduction}

Monitoring structural integrity, maintaining planned functionality, and managing the operation of critical structural systems represent paramount challenges in modern engineering~\cite{ML_perspective,Avci_review,Torzoni_DML}. These systems are often threatened by material deterioration, cyclic loading, and extreme events. Such variability, combined with natural hazards increasingly driven by climate change~\cite{committee2018climate}, calls for a new level of ``adaptivity'' in the monitoring and maintenance of engineering systems~\cite{MT_AIF}. This is essential to respond effectively to changing conditions over time and ultimately to minimize the risks associated with compromised structures~\cite{farrar2007introduction, brownjohn2007structural,infrastructure2021comprehensive}. Additionally, asset managers must ensure the practical and effective use of services, avoiding unforeseen access limitations that could lead to higher costs for users~\cite{thoft2009life,Hudson_River}. Decision-makers must also maximize profits (or minimize losses), which results into a multi-objective optimization problem under uncertainty. This challenge can be effectively addressed by combining structural health monitoring (SHM) with sequential decision-making within a digital twin (DT) framework, enabling effective damage identification strategies and optimal action-planning schedules.

A computational model that overlooks the uncertain and dynamic nature of engineering systems limits its personalization, which in turn reduces the model accuracy and the value of insights derived from it~\cite{moya2020physically,cherifi2022simulations,arcones2024modelbias}. According to the 2024 report by the \textit{National Academies of Engineering, Science, and Medicine}~\cite{Foundational}, a DT can be distinguished from instances of digital models (forward models from parameters to outputs of interest) or digital shadows (data assimilation and model updating engines)~\cite{KRITZINGER20181016}. A DT is defined as a personalized virtual representation of specific attributes of a natural or engineered system or process. It relies on computational models that dynamically mirror the physical counterpart by continuously assimilating sensor data and refining predictive capabilities accordingly. This continuous updating enables the simulation of what-if scenarios, feeding predictive decision-making that is tailored to realize value. Rooted in the ``Dynamic Data Driven Applications Systems'' paradigm~\cite{darema2004dynamic}, this bi-directional interaction between the virtual and the physical forms a feedback loop that propels the DT concept toward autonomy rather than simple automation. DT applications today span a variety of fields, including SHM and predictive maintenance~\cite{Torzoni_DT,li2017dynamic,review_2,karve2020digital, hfdt2024}, additive manufacturing~\cite{phua2022digital}, smart cities~\cite{review_6}, urban sustainability~\cite{tzachor2022potential}, railway systems management~\cite{arcieri_pomdp,arcieri_pomdp2}, spacecraft operations~\cite{henaogarcia2025scitech}, and aerial vehicles monitoring and control~\cite{mcclellan2022physics,Sisson_2,pgm_wilcox_dt,kapteyn2020data, tezzele2024adaptive}, among others. Beyond engineering, the DT paradigm is also being deployed in healthcare for advancing medical diagnosis and personalized treatment~\cite{corral2020digital,chaudhuri2023predictive}, as well as in climate science~\cite{bauer2021digital}.

Adopting a DT perspective for SHM purposes is crucial to enable condition-based or predictive maintenance practices~\cite{Glaser,Achenbach,tesi_Matteo}. To this aim, in-situ inspections are inadequate for continuous and global monitoring. On the other hand, by assimilating sensor data from permanent data collecting systems~\cite{Torzoni_DML,Springer_Ubertini}, vibration-based techniques enable automated damage identification and evolution tracking. This paradigm holds the potential to unlock personalized monitoring and management programs~\cite{DT_review}, yielding numerous benefits throughout the system life cycle, including more informed assessments of structural safety, better resource allocation, and increased system availability.

\begin{figure}[ht]
\centering
\includegraphics[width=1\linewidth]{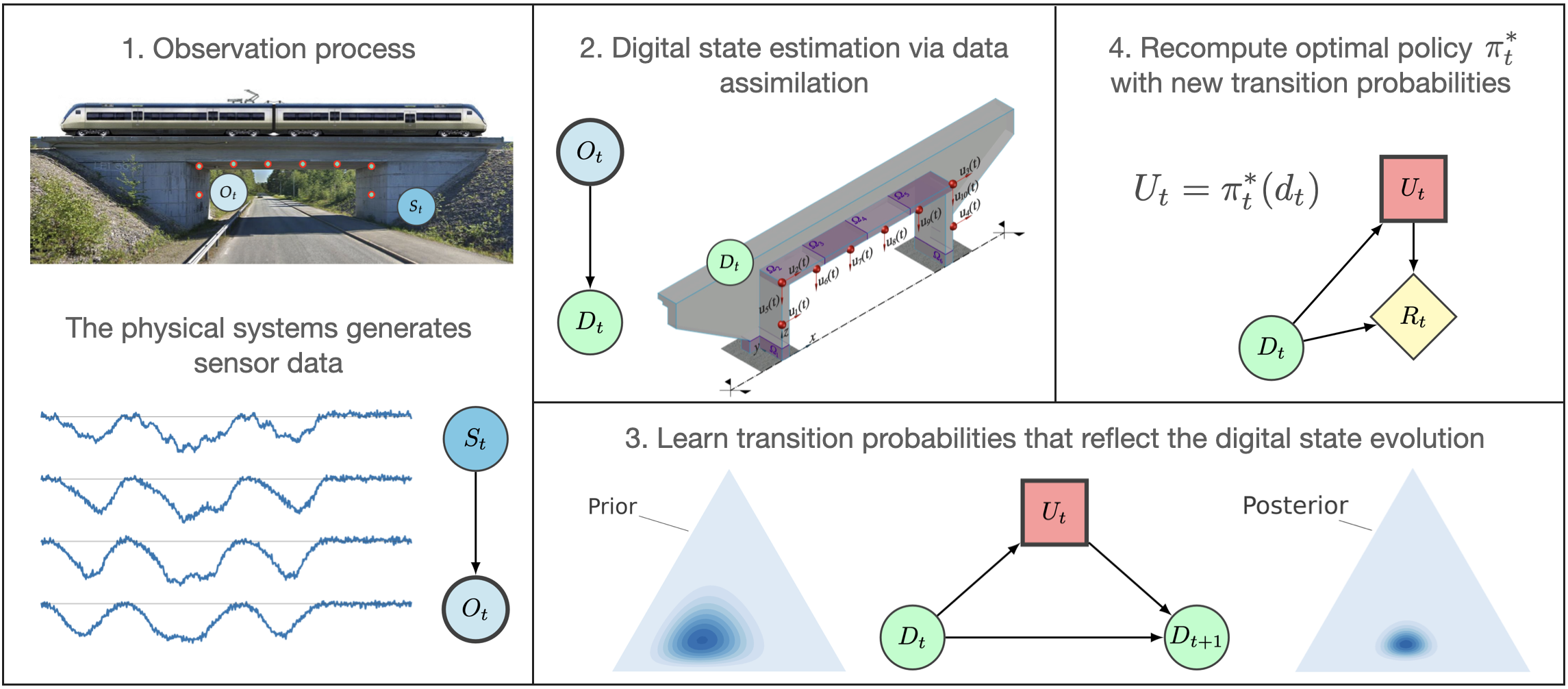}
\caption{Graphical abstraction of the end-to-end information flow enabled by the probabilistic graphical model. Capital letters denote time-dependent random variables, while the corresponding lowercase letters refer to their specific realizations. Circular nodes denote random variables, square nodes denote actions, and diamond nodes denote the objective function. Nodes with bold outlines denote observed quantities, while nodes with thin outlines denote unobserved variables to be inferred. Directed edges encode conditional dependencies between variables. (Panel 1) The hidden physical state $S_t\sim p(s_t)$ governs the emission of observational data $O_t \sim p(o_t)$. (Panel 2) The assimilation of observational data characterizes the physical-to-digital information flow through the inference of a digital state $D_t \sim p(d_t)$, which captures the essential features of the physical state. (Panel 3) The update of the digital state occurs within a sequential Bayesian inference formulation, where a stochastic parametrization of the state transition probabilities enables online learning through continuous updating of state transition beliefs. (Panel 4) The updated digital state informs the digital-to-physical information flow by guiding the selection of control actions $U_t\sim p(u_t)$ to maximize the performance of the asset-twin system, as measured by the reward $R_t\sim p(r_t)$. The optimal control policy $\pi^*_t$ governing this process is continuously refined as the transition beliefs are updated.}  
\label{fig:overview}
\end{figure}

In~\cite{Torzoni_DT}, the authors presented a DT framework for engineering structures based on probabilistic graphical models (PGMs) for predictive DTs introduced in~\cite{pgm_wilcox_dt}. Specifically, the DT is built on a dynamic Bayesian network (DBN) with decision nodes~\cite{koller2009probabilistic,2010artificial}, which encodes the observation-to-decision interaction and its temporal evolution. The computational flow spans from physical to digital through data assimilation and inference, and then back to the asset via control actions, while also accounting for uncertainty quantification and propagation. As depicted in panel 1 of Figure~\ref{fig:overview}, vibration recordings are assimilated for structural health diagnostics (physical-to-digital information flow), leveraging the flexibility of deep learning (DL) models~\cite{5_TMM,tong2024invaert,franco2023latent,Torzoni_MF,vlachas2024reduced,Torzoni_EWSHM}. These DL models are preliminarily trained offline on structural health scenarios generated from a numerical model that simulates the structural response to applied loads. Once trained, the DL predictors map vibration recordings to parameters describing the presence, location, and severity of damage. This initial estimation is integrated into a sequential Bayesian inference formulation to continually update a digital state encoding the health condition of the asset (see panel 2 in Figure~\ref{fig:overview}). This updated characterization is then used to forecast future damage evolution via DBN unrolling. The predicted trajectories guide a Markov decision process (MDP) for maintenance and management actions (digital-to-physical information flow). The health-aware control policy is also precomputed offline using dynamic programming~\cite{bellman1957} to solve the planning problem induced by the PGM. After an action is taken, the system state evolves, and the perception-action cycle begins~again.

This work enhances the aforementioned framework by incorporating adaptivity to improve decision-making capabilities. Specifically, the reference PGM includes a transition model that serves as a control-dependent predictor, propagating digital state beliefs forward in time. While this internal model is typically predefined using historical data or physics-based models, as in our previous work~\cite{Torzoni_DT}, here we adopt a stochastic parametrization that enables online learning of state transition beliefs through Bayesian updates (see panel 3 in Figure~\ref{fig:overview}). This adaptivity ensures that the DT remains synchronized with its physical counterpart and can effectively respond to changing conditions over time. Moreover, moving beyond precomputed optimal policies --- based on the unrealistic assumption of precisely known transition dynamics --- we formulate a parametric MDP and develop dynamic policies with continuous precision updates (see panel 4 in Figure~\ref{fig:overview}). Specifically, we leverage the refined transition dynamics and employ model-based reinforcement learning (MBRL) to optimize control policies over the expected remaining lifespan of the structure. A similar approach, not in a DT context, for partially observable MDP was proposed in~\cite{Memarzadeh2015} where the authors utilize a Bayes-adaptive framework to address real-time decision-making for the management of wind farms under uncertainty.

The paper is organized as follows. Section~\ref{sec:math} reviews the components of the DT framework as presented in~\cite{Torzoni_DT}. Section~\ref{sec:methodology} describes the proposed adaptive capabilities within the context of parametric MDPs, extending~\cite{tezzele2024adaptive} to a broader class of probability distributions. This section also shows how optimal policies can be systematically updated using RL techniques. Section~\ref{sec:res} assesses the computational procedure through the simulated monitoring of a railway bridge. Conclusions are drawn in Section~\ref{sec:conclusion}.

\section{Probabilistic graphical models for digital twins}
\label{sec:math}
Dynamic Bayesian networks are a specific type of PGM used to model probability distributions that evolve over time, making them suitable for inference and learning in dynamic environments~\cite{koller2009probabilistic,2010artificial}. This section introduces the asset-twin DBN model used in~\cite{Torzoni_DT}, which is based on the mathematical abstraction proposed in~\cite{pgm_wilcox_dt}. We begin by introducing DBNs, and then detail the specific model used in this work.

\subsection{Dynamic Bayesian networks}
Probabilistic graphical models represent joint probability distributions through efficient factorization, leveraging the conditional independence among random variables induced by the graph structure. Bayesian networks are a specific type of PGMs represented by directed acyclic graphs. A directed graph $\mathcal{G}=\langle \mathcal{V}, \mathcal{E}\rangle$ is a graph composed by a set of vertices $\mathcal{V}$ connected by directed edges in $\mathcal{E} \subseteq \lbrace \mathcal{V} \times \mathcal{V}  \rbrace$ which is a set of ordered pairs of vertices. 

In a Bayesian network, $\mathcal{V}$ represents a set of random variables, and the edges connecting the nodes represent causal relationships through conditional probability distributions (CPDs). Furthermore, given a set $\mathcal{X}=\lbrace X_1,\ldots,X_N\rbrace$ of $N$ random variables (whether observed or hidden), we can factorize their joint probability distribution using the chain rule of probability, effectively leveraging the conditional independence structure induced by the graph topology. To this end, each node $X_i$, $i=1,\ldots,N$, has an associated factor $\phi^i=p(X_i\mid \text{Pa}_{X_i})$, which quantifies the influence of its parent nodes $\text{Pa}_{X_i}$, reflecting the CPDs implied by the joint distribution. Nodes with no parents represent independent variables, specified by a prior (unconditional) distribution.
 
To extend Bayesian networks to settings where variables evolve over time, we denote the template variable $X_i\in\mathcal{X}$, $i=1,\ldots,N$, at a specific point in time $t$ as $X_i^{(t)}$. The trajectories of the (dynamic) Bayesian network refer to the complete set of possible realizations over a time horizon $[0,T]$, with $t\in[0,T]$. The joint distribution for all possible trajectories reads as follows:
\begin{equation}
p(\mathbf{X}^{(t_1 : \,t_2)}) = p(\mathbf{X}^{(t_1)},\ldots, \mathbf{X}^{(t_2)}),
\label{eq:full_joint}
\end{equation}
where $\mathbf{X}=(X_1,\ldots,X_N)^\top$, and $\mathbf{X}^{(t_1:\,t_2)}$ denotes the set of variables $\lbrace\mathbf{X}^{(t)}: t \in [t_1,t_2]\rbrace$, $t_1 < t_2$. 

To alleviate the computational burden associated with representing the joint distribution in Equation~\eqref{eq:full_joint}, a DBN typically relies on a set of simplifying assumptions. First, we consider a time discretization, which involves partitioning the time horizon $[0,T]$ into time slices $t = 0,\ldots, T$, with an arbitrary step $\Delta t$. In this work, we assume regularly spaced time steps with granularity equal to the time between two consecutive observations of the system. Second, we assume that the conditional dependencies between consecutive random variables are represented by a Markov process. Accordingly, each variable $X_i^{(t)}$ can have parents only in the current or immediately preceding time slice. This conditional independence structure allows us to compactly write $p(\mathbf{X}^{(0\,:\,T)})$ as
\begin{equation}
    p(\mathbf{X}^{(0\,:\,T)}) = p(\mathbf{X}^{(0)}) \prod_{t=0}^{T-1}p(\mathbf{X}^{(t+1)} \mid  \mathbf{X}^{(t)}).
\label{eq:markov}
\end{equation}
Hence, the probability distribution over different trajectories is fully characterized by the initial state distribution $p(\mathbf{X}^{(0)})$ and the transition model $p(\mathbf{X}^{(t+1)}\mid \mathbf{X}^{(t)})$. This implies that probabilities at time $t+1$ are conditionally independent of all states prior to time $t$, given the state at time $t$.

Typically, a third assumption concerns the stationarity of the underlying Markov process, which allows the conditional dependencies between consecutive random variables to be represented by an invariant transition model:
\begin{equation}
    p(\mathbf{X}^{(t+1)}\mid \mathbf{X}^{(t)}) = p(\mathbf{X}'\mid \mathbf{X}) \quad \forall t, \quad \mathbf{X},\mathbf{X}' \subseteq \mathcal{X}.
\label{eq:stationarity}
\end{equation}
As detailed in the following, in this work we relax this assumption, resulting in adaptive DTs capable of tracking systems that evolve according to a non-stationary process.

The transition model in Equation~\eqref{eq:stationarity} can be represented by a 2-time-slice Bayesian network (2-TBN) for a process over $\mathcal{X}$, which encodes the following CPD:
    \begin{equation}
    p(\mathbf{X}'\mid \mathbf{X}) = \prod_{i=1}^N p(X_i'\mid \text{Pa}_{X_i'}), 
    \end{equation}
where, in addition to $\mathbf{X}'$, the 2-TBN includes only those variables in $\mathbf{X}$ with a direct effect on $\mathbf{X}'$. 
 
Given a Bayesian network $\mathcal{B}_0$ representing the initial distribution $p(\mathbf{X}^{(0)})$, a DBN over the set of template variables $\mathcal{X}$ is defined by a pair $\langle\mathcal{B}_0,\mathcal{B}_\to\rangle$, where $\mathcal{B}_\to$ is a 2-TBN for the process over $\mathcal{X}$. The graph topology and conditional dependencies for $t=0$ follow those in $\mathcal{B}_0$, while for $t>0$ they follow those in $\mathcal{B}_\to$. The resulting DBN provides a single compact model that captures the system dynamics and yields distributions over different trajectories. It can be interpreted as a sequence of Bayesian networks, one for each time slice.

For any desired time horizon $T>0$, the distribution over $\mathbf{X}^{(0\,:\,T)}$ is obtained by unrolling the DBN until $T$, repeating the core structure of the graph at each time step. Given a sequence of observations, a DBN can be unrolled until it fully encompasses the observed data. Additional time slices beyond the last observation do not affect inference within the observation period. Once unrolled, the DBN becomes a standard Bayesian network, and beliefs can be updated and propagated as for static Bayesian networks.
  
\subsection{Probabilistic graphical model for predictive structural digital twins} \label{subsec:used_dbn}

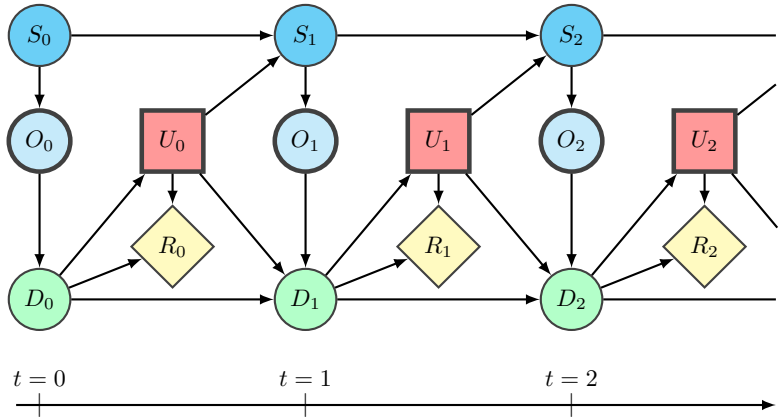
\begin{figure}[ht]
\center
\begin{tikzpicture}[scale=.7, every node/.style={scale=.9}]

\node [P_node] (P_0) at (0,5) {};
\node [] () at (0,5) {$S_{0}$};
\node [D_node] (D_0) at (0,0) {};
\node [] at (0,0) {$D_{0}$};
\node [O_node] (O_0) at (0,3) {};
\node [] at (0,3) {$O_{0}$};
\node [U_node_act] (U_A_0) at (2.5,3) {};
\node [] at (2.5,3) {$U_{0}$};
\node [R_node] (R_0) at (2.5,1) {};
\node [] at (2.5,1) {$R_{0}$};

\node [P_node] (P_1) at (5,5) {};
\node [] () at (5,5) {$S_{1}$};
\node [D_node] (D_1) at (5,0) {};
\node [] at (5,0) {$D_{1}$};
\node [O_node] (O_1) at (5,3) {};
\node [] at (5,3) {$O_{1}$};
\node [U_node_act] (U_A_1) at (7.5,3) {};
\node [] at (7.5,3) {$U_{1}$};
\node [R_node] (R_1) at (7.5,1) {};
\node [] at (7.5,1) {$R_{1}$};

\node [P_node] (P_2) at (10,5) {};
\node [] () at (10,5) {$S_{2}$};
\node [D_node] (D_2) at (10,0) {};
\node [] at (10,0) {$D_{2}$};
\node [O_node] (O_2) at (10,3) {};
\node [] at (10,3) {$O_{2}$};
\node [U_node_act] (U_A_2) at (12.5,3) {};
\node [] at (12.5,3) {$U_{2}$};
\node [R_node] (R_2) at (12.5,1) {};
\node [] at (12.5,1) {$R_{2}$};

\node [] (t0) at (-0.6,-2) {};
\node [] () at (0,-2) {$|$};
\node [] () at (0,-1.5) {$t=0$};
\node [] () at (5,-2) {$|$};
\node [] () at (5,-1.5) {$t=1$};
\node [] () at (10,-2) {$|$};
\node [] () at (10,-1.5) {$t=2$};
\node [] (t2) at (14,-2) {};

\node [] (P_E) at (14,5) {};
\node [] (D_E) at (14,0) {};
\node [] (U_E_P) at (14,4.2) {};
\node [] (U_E_D) at (14,1.2) {};

\draw[-latex,thick,black] (P_0) to (P_1);
\draw[-latex,thick,black] (P_1) to (P_2);

\draw[-latex,thick,black] (P_0) to (O_0);
\draw[-latex,thick,black] (O_0) to (D_0);
\draw[-latex,thick,black] (D_0) to (U_A_0);
\draw[-latex,thick,black] (U_A_0) to (R_0);
\draw[-latex,thick,black] (D_0) to (R_0);

\draw[-latex,thick,black] (D_0) to (D_1);
\draw[-latex,thick,black] (U_A_0) to (P_1);
\draw[-latex,thick,black] (P_1) to (O_1);
\draw[-latex,thick,black] (O_1) to (D_1);
\draw[-latex,thick,black] (U_A_0) to (D_1);
\draw[-latex,thick,black] (D_1) to (U_A_1);
\draw[-latex,thick,black] (U_A_1) to (R_1);
\draw[-latex,thick,black] (D_1) to (R_1);

\draw[-latex,thick,black] (D_1) to (D_2);
\draw[-latex,thick,black] (U_A_1) to (P_2);
\draw[-latex,thick,black] (P_2) to (O_2);
\draw[-latex,thick,black] (O_2) to (D_2);
\draw[-latex,thick,black] (U_A_1) to (D_2);
\draw[-latex,thick,black] (D_2) to (U_A_2);
\draw[-latex,thick,black] (U_A_2) to (R_2);
\draw[-latex,thick,black] (D_2) to (R_2);

\draw[-latex,thick,black] (t0) to (t2);

\draw[-,thick,black] (D_2) to (D_E);
\draw[-,thick,black] (P_2) to (P_E);
\draw[-,thick,black] (U_A_2) to (U_E_P);
\draw[-,thick,black] (U_A_2) to (U_E_D);

\end{tikzpicture}
\caption{Dynamic Bayesian network encoding the asset-twin dynamical system, unrolled over two time slices. Circle nodes denote random variables, square nodes denote actions, and diamond nodes denote the objective function. Nodes with bold outlines denote observed quantities, while nodes with thin outlines denote unobserved quantities to be inferred. Directed edges represent the variables' dependencies encoded via conditional probability distributions.}
\label{fig:unrolled_dbn}
\end{figure}

The PGM used to represent the dynamic interaction between the physical and virtual domains is depicted in Figure~\ref{fig:unrolled_dbn} as a dynamic decision network, i.e., a DBN with decision nodes. In the graphical representation, circle nodes denote random variables, square nodes indicate actionable inputs, and diamond-shaped nodes symbolize the objective function, all at discrete time steps. Each time the DT is updated by assimilating observational data, the PGM advances of one time step, with $t\in\lbrace0,\ldots,T\rbrace$. Here, $t=0$ marks the time at which the DT enters operation, and $t=T$ represents its lifetime horizon. Nodes with bold outlines represent observed quantities, while nodes with thin outlines represent unobserved quantities to be inferred. The PGM is sparsely connected, with edges encoding known or assumed conditional dependencies between variables.

We use capital letters to denote random variables representing the quantities involved in our abstraction, with subscripts indicating the time index. Corresponding lowercase letters represent realizations of these random variables. Calligraphic script variables denote the space of possible values that these quantities can take. For instance, the unknown physical state, which reflects the ground truth variability of interest, is denoted as $S_t\sim p(s_t)$, where $s_t$ represents the realization of the random variable $S_t$ at time $t$, and $p(s_t)$ represents the probability that $S_t=s_t$.
 
The digital state $D_t \sim p(d_t)$ aims to capture the variability in the physical state. It may include various types of information, such as initial conditions, boundary conditions, material properties, and other relevant quantities describing the asset (or process) of interest. However, a DT does not need to be an exact virtual replica to provide maximum value; rather, it should capture the essential features and variations relevant for diagnosis, prediction, and decision-making~\cite{Ferrari2024}. Although there are no strict constraints in this regard, we define the digital state as a categorical random variable that can assume values among a set of predefined configurations in the digital state space $\mathcal{D}$.

The physical-to-digital information flow from $S_t$ to $D_t$ is governed by the continuous collection and assimilation of observational data \mbox{$O_t \sim p(o_t)$}. These may include sensor recordings, inspection data, or diagnostic information. Since the physical state $S_t$ is only partially and indirectly observable, the digital state $D_t$ cannot be uniquely determined. Instead, it provides evidence for digital state configurations that could be consistent with the given observations.

The updated digital state $D_t$ informs the digital-to-physical information flow, i.e., the control actions. In Figure~\ref{fig:unrolled_dbn}, $U_t\sim p(u_t)$ is a decision variable representing the action to be taken. The space of actions $\mathcal{U}$ may include interventions affecting the physical state, adjustments to the operational conditions, or modifications to the observational process. Each action is characterized by its own transition dynamics across the system's state space.

Finally, the reward node $R_t\sim p(r_t)$ measures the performance of the asset-twin system. Specifically, the reward function provides values within a reward space $\mathcal{R}$, measuring the expected utility of DBN trajectories that we maximize when solving the planning problem induced by the PGM.
\begin{figure}[ht]
\centering
\captionsetup[subfigure]{justification=centering}\subfloat{
\begin{tikzpicture}[scale=.8, every node/.style={scale=.85}]
\node [P_node] (P_0) at (0,5) {};
\node [] () at (0,5) {$S_{0}$};
\node [D_node] (D_0) at (0,0) {};
\node [] at (0,0) {$D_{0}$};
\node [O_node] (O_0) at (0,3) {};
\node [] at (0,3) {$O_{0}$};
\node [U_node_act] (U_A_0) at (2.5,3) {};
\node [] at (2.5,3) {$U_{0}$};
\node [R_node] (R_0) at (2.5,1) {};
\node [] at (2.5,1) {$R_{0}$};
\node[] at (0,-1.5) {\textcolor{white}{T}};

\node[] at (1.25,6.4) {$\mathcal{B}_0$};

\draw[-latex,thick,black] (P_0) to (O_0);
\draw[-latex,thick,black] (O_0) to (D_0);
\draw[-latex,thick,black] (D_0) to (U_A_0);
\draw[-latex,thick,black] (U_A_0) to (R_0);
\draw[-latex,thick,black] (D_0) to (R_0);
\end{tikzpicture}}\hspace{1cm}\subfloat{\begin{tikzpicture}[scale=.8, every node/.style={scale=.85}]

\begin{pgfonlayer}{background}
    \fill[fill=gray, fill opacity=0.1, draw=black, dashed, rounded corners] (-0.85, -1) rectangle (3.4, 6);
    \fill[fill=gray, fill opacity=0.1, draw=black, dashed, rounded corners] (4.15, -1) rectangle (8.4, 6);
\end{pgfonlayer}
\node[] at (1.275,-1.5) {Time slice $t-1$};
\node[] at (6.275,-1.5) {Time slice $t$};
\node[] at (3.775,6.4) {$\mathcal{B}_\to$};

\node [P_node] (P_0) at (0,5) {};
\node [] () at (0,5) {$S_{t-1}$};
\node [D_node] (D_0) at (0,0) {};
\node [] at (0,0) {$D_{t-1}$};
\node [U_node_act] (U_A_0) at (2.1,3) {};
\node [] at (2.1,3) {$U_{t-1}$};

\node [P_node] (P_1) at (5,5) {};
\node [] () at (5,5) {$S_{t}$};
\node [D_node] (D_1) at (5,0) {};
\node [] at (5,0) {$D_{t}$};
\node [O_node] (O_1) at (5,3) {};
\node [] at (5,3) {$O_{t}$};
\node [U_node_act] (U_A_1) at (7.5,3) {};
\node [] at (7.5,3) {$U_{t}$};
\node [R_node] (R_1) at (7.5,1) {};
\node [] at (7.5,1) {$R_{t}$};

\draw[-latex,thick,black] (P_0) to (P_1);
\draw[-latex,thick,black] (D_0) to (D_1);
\draw[-latex,thick,black] (U_A_0.east) to (P_1);
\draw[-latex,thick,black] (P_1) to (O_1);
\draw[-latex,thick,black] (O_1) to (D_1);
\draw[-latex,thick,black] (U_A_0.east) to (D_1);
\draw[-latex,thick,black] (D_1) to (U_A_1);
\draw[-latex,thick,black] (U_A_1) to (R_1);
\draw[-latex,thick,black] (D_1) to (R_1);
\end{tikzpicture}}
\caption{Dynamic Bayesian network encoding the asset-twin dynamical system: Bayesian network representing the initial distribution $\mathcal{B}_0$; two-time-slice Bayesian network for the process $\mathcal{B}_\to$. Circle nodes denote random variables, square nodes denote actions, and diamond nodes denote the objective function. Nodes with bold outlines denote observed quantities, while nodes with thin outlines denote unobserved quantities to be inferred. Directed edges represent the variables' dependencies encoded via conditional probability distributions.
\label{fig:B0_Bt}}
\end{figure}
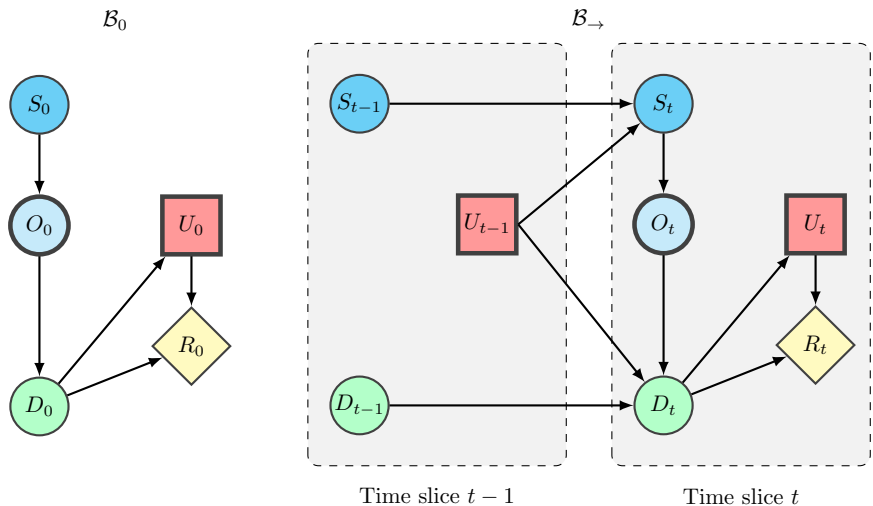

Since we assume discrete spaces for digital states and control actions, the related causal relationships are modeled using conditional probability tables (CPTs). In a CPT, each column contains the CPD of a node's value given the values of its parent nodes. Each column sums to one, as the entries represent an exhaustive set of possible outcomes for the variable. A node with no parents has only one column, indicating the prior probabilities of each possible value. CPTs can be viewed as a set of parameters representing prior knowledge about random variables. Thanks to the discrete DBN setting, we can propagate and update beliefs exactly with a single pass of the sum-product message-passing algorithm~\cite{Pearl1982}.

The Bayesian network $\mathcal{B}_0$ and the 2-TBN $\mathcal{B}_\to$, jointly forming our PGM, are shown in Figure~\ref{fig:B0_Bt}. $\mathcal{B}_0$ encodes the initial distribution of the random variables, which can be derived from expert knowledge to assign a prior distribution to the initial state of the asset-twin system. In contrast, $\mathcal{B}_\to$ encodes the inter-slice conditional dependencies, modeling the temporal relationships between the random variables. The PGM enables the extraction of the joint distribution of the network's trajectories, conditioned on observed data $O_t=o_t$ and enacted actions $U_t=u_t$. Specifically, the joint belief state can be factorized from the initial time step $t = 0$ up to the current time $t_c$, using the following sequential Bayesian inference formulation:
\begin{equation}
p(D_{0:t_c}, U_{0:t_c}, R_{0:t_c}\mid o_{0:t_c},u_{0:t_c})\propto \prod_{t=0}^{t_c}\Bigl[\phi_t^\text{update} \phi_t^\text{control} \phi_t^\text{reward} \Bigr],
\label{eq:joint_factor}
\end{equation}
with factors:
\begin{align}
    &\phi_t^\text{update} = p(D_t \mid  D_{t-1}, U_{t-1} = u_{t-1},O_{t} = o_{t}),\\
    &\phi_t^\text{control} =p(U_t\mid D_t),\\    
    &\phi_t^\text{reward} = p(R_t \mid  D_t,U_t =u_t).
\end{align}

The factor $\phi_t^{\text{update}}$ updates the digital state belief $D_t$ to reflect the (hidden) physical state $S_t$. First, it accounts for a control-dependent transition model, describing the expected evolution of the digital state given the previous belief $D_{t-1}$ and the most recent action taken $U_{t-1} =u_{t-1}$. Additionally, it integrates this prior with the data assimilation outcome by conditioning on $O_{t}=o_{t}$. We can explicitly split the contributions of transition dynamics and data assimilation by further factorizing $\phi_t^{\text{update}}$ into a predictor-corrector form update, as follows\footnote{The causal structure required for this factorization is not included in the set of conditional dependencies induced by the graph depicted in Figure~\ref{fig:unrolled_dbn}. Graphically, this would require an arrow from $D_t$ to $O_{t}$, but in our model, the arrow is depicted in the opposite direction to reflect the information flow within the digital twin, consistent with~\cite{Torzoni_DT,pgm_wilcox_dt}. For further details, the reader may refer to~\cite{koller2009probabilistic}.}:
\begin{equation}
\phi_t^\text{update} \propto \phi_t^\text{data} \phi_t^\text{history},
\label{eq:fact_update}
\end{equation}
where:
\begin{align}
    &\phi_t^{\text{history}}=p(D_t \mid  D_{t-1}, U_{t-1} = u_{t-1}),\\
    &\phi_t^\text{data} = p(O_{t} = o_{t} \mid  D_t).
\end{align} 
The transition model $\phi_t^{\text{history}} : \mathcal{D} \times \mathcal{D} \times \mathcal{U} \mapsto [0,1]$ acts as a forward-time predictor. It can be visualized as a CPT describing the expected transition dynamics from $D_{t-1}$ to $D_t$ for each possible action $U_{t-1} =u_{t-1}$. This CPT should incorporate any prior knowledge the DT designer has regarding the asset and relevant operational conditions. It can be derived offline from historical data and accumulated experience~\cite{state_trans_1,state_trans_2,Bianchi_ann}, as well as from physics-based or empirical evolution models~\cite{biondini_review,biondini2017time,Gabriel_2}. In this paper, we introduce a stochastic parametrization of $\phi_t^{\text{history}}$, similar to~\cite{tezzele2024adaptive}, enabling online learning of state transition beliefs via Bayesian updates, as described in Section~\ref{subsec:online_learning}. In contrast, the factor $\phi_t^{\text{data}}$ is associated with the assimilation of observational data $O_{t} = o_{t}$, as further described at the end of this section.

The factor $\phi_t^{\text{control}} : \mathcal{U} \times \mathcal{D} \mapsto [0,1]$ evaluates the belief about the next optimal control action. This factor corresponds to a CPT encoding a control policy $\pi: \mathcal{D} \mapsto \mathcal{U}$. The actual action to be taken is selected as the best point estimate $u_t=\argmax_{u\in\mathcal{U}}\pi(D_t)$, i.e., the conversion from probabilistic to deterministic control. The computation of $\pi$ is described in Section~\ref{subsec:dyn_poli}.

Finally, $\phi^{\text{reward}} : \mathcal{D} \times \mathcal{U} \to \mathcal{R}$ encapsulates the evaluation of the objective function. The two edges leading to $R_t$ represent a reward function that quantifies the costs associated with the digital state and control actions, respectively. 
  
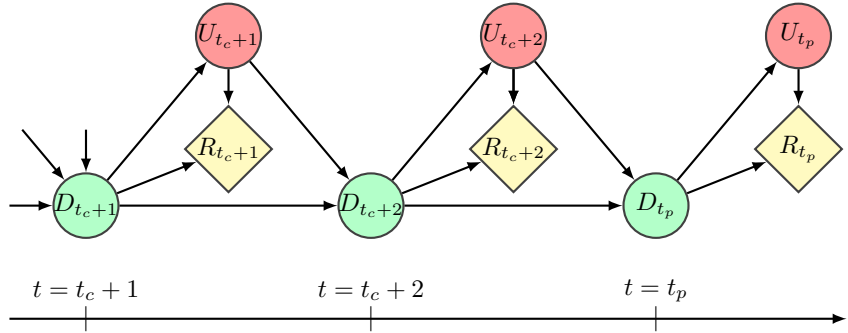
\begin{figure}[ht]
\center
\begin{tikzpicture}[scale=.75, every node/.style={scale=.95}]

\node [D_node] (D_0) at (0,0) {};
\node [] at (0,-0) {$D_{t_c+1}$};
\node [U_node_prob] (U_P_0) at (2.5,3) {};
\node [] at (2.5,3) {$U_{t_c+1}$};
\node [R_node] (R_0) at (2.5,1) {};
\node [] at (2.5,1) {$R_{t_c+1}$};

\node [D_node] (D_1) at (5,-0) {};
\node [] at (5,0) {$D_{t_c+2}$};
\node [U_node_prob] (U_P_1) at (7.5,3) {};
\node [] at (7.5,3) {$U_{t_c+2}$};
\node [R_node] (R_1) at (7.5,1) {};
\node [] at (7.5,1) {$R_{t_c+2}$};

\node [D_node] (D_2) at (10,0) {};
\node [] at (10,0) {$D_{t_p}$};
\node [U_node_prob] (U_P_2) at (12.5,3) {};
\node [] at (12.5,3) {$U_{t_p}$};
\node [R_node] (R_2) at (12.5,1) {};
\node [] at (12.5,1) {$R_{t_p}$};

\node [] (D_E) at (-1.5,0) {};
\node [] (O_E) at (0,1.5) {};
\node [] (U_E) at (-1.25,1.5) {};

\node [] (t0) at (-1.5,-2) {};
\node [] () at (0,-2) {$|$};
\node [] () at (0,-1.5) {$t=t_c+1$};

\node [] () at (5,-2) {$|$};
\node [] () at (5,-1.5) {$t=t_c+2$};

\node [] () at (10,-2) {$|$};
\node [] () at (10,-1.5) {$t= t_p$};

\node [] (t3) at (13.5,-2) {};

\draw[-latex,thick,black] (D_0) to (U_P_0);
\draw[-latex,thick,thick,black] (U_P_0) to (D_1);
\draw[-latex,thick,black] (U_P_0) to (R_0);
\draw[-latex,thick,black] (D_0) to (R_0);

\draw[-latex,thick,black] (D_0) to (D_1);
\draw[-latex,thick,black] (D_1) to (U_P_1);
\draw[-latex,thick,black] (U_P_1) to (R_1);
\draw[-latex,thick,black] (D_1) to (R_1);

\draw[-latex,thick,black] (D_1) to (D_2);
\draw[-latex,thick,black] (D_2) to (U_P_2);
\draw[-latex,thick,thick,black] (U_P_1) to (D_2);
\draw[-latex,thick,black] (U_P_1) to (R_1);
\draw[-latex,thick,black] (D_2) to (R_2);

\draw[-latex,thick,black] (U_P_2) to (R_2);

\draw[-latex,thick,black] (t0) to (t3);

\draw[-latex,thick,black] (D_E) to (D_0);
\draw[-latex,thick,black] (O_E) to (D_0);
\draw[-latex,thick,black] (U_E) to (D_0);

\end{tikzpicture}
\caption{Dynamic Bayesian network used to predict the future evolution of digital states and control actions. Circle nodes denote random variables, and diamond nodes denote the objective function. Directed edges represent the variables' dependencies encoded via conditional probability distributions.}
\label{fig:unrolled_dbn_prediction}
\end{figure}

Given the current digital state $D_{t_c}$, probabilistic estimates of future digital states and control actions are made by unrolling a simplified DBN up to a prediction time $t_p > t_c$. This prediction DBN, depicted in Figure~\ref{fig:unrolled_dbn_prediction}, retains a structure similar to the one discussed above but is stripped of the nodes related to observations, data assimilation, and action execution. At prediction stage, information from sensing and performed actions is no longer available. Accordingly, the actions $U_t$ are depicted as random variables rather than decision nodes, as they represent predictions of what-if scenarios beyond data assimilation, made through uncertainty propagation via DBN unrolling. As a result, the factorization over the trajectories in Equation~\eqref{eq:joint_factor} can be extended over the prediction time steps $t = t_c + 1,\ldots,t_p$ by augmenting the belief state as follows:
\begin{equation}
p(D_{t_c+1:t_p}, U_{t_c+1:t_p}, R_{t_c+1:t_p} \mid  D_{t_c},U_{t_c})
= \prod_{t=t_c+1}^{t_p}{[\phi_t^\text{history}  \phi_t^\text{control} \phi_t^\text{reward}]}.\label{eq:joint_factor_prediction}
\end{equation}
In this case, the factors $\phi_t^\text{history} = p(D_t\mid D_{t-1}, U_{t-1})$ and $\phi_t^\text{reward} = p(R_t\mid  D_t, U_t)$ enable future state predictions and probabilistic reward estimates, respectively, conditioned on predicted control actions that have not yet been enacted. As clarified in Section~\ref{subsec:dyn_poli}, factors $\phi_t^\text{history}$ and $\phi_t^\text{control}$ actually encode dynamically updated transition models and control policies, respectively. Indeed, after updating the transition model $\phi_t^\text{history}$ with the evidence collected up to $t_c$, we also recompute an optimal policy $\pi_{t_c}$, such that $U_{t_c:t_p}\sim\pi_{t_c}(D_{t_c:t_p})$.

In this work, the asset of interest is a deteriorating structural system, although the framework outlined above can be adapted to other types of systems or applications beyond SHM. The asset is dynamic, meaning its state evolves over time according to laws that depend on both the system itself and external factors. These external factors can either improve or degrade the condition of the asset and influence its deterioration rate. Such factors may include long-term degradation caused by chemical, physical, or mechanical processes, as well as sudden changes due to discrete damage events or interventions by external agents~\cite{zakic1991classification}.

We simulate the monitored asset using a physics-based computational model. Specifically, the structure is modeled as a linear-elastic continuum, and its dynamic response to applied loads is described through elasto-dynamics equations, assuming linearized kinematics. The model is spatially discretized using finite elements, and its solution is advanced over time to produce synthetic observational data. This computational model serves as a data generator, controlled by a vector of $N_\text{par}$ parameters $\boldsymbol{\mu} \in\mathbb{R}^{N_\text{par}}$, which describe the operational, damage, and (possibly) environmental conditions. Further details on the modeling choices for the H{\"o}rnefors railway bridge case study considered in this work will be provided in Section~\ref{sec:res}.

The observational data consist of vibration recordings shaped as multivariate time series\linebreak $\mathbf{U}(\boldsymbol{\mu})=[\mathbf{u}_1(\boldsymbol{\mu}),\ldots,\mathbf{u}_{N_s}(\boldsymbol{\mu})]\in\mathbb{R}^{L\times N_s}$, made up of $N_s$ time series, each containing $L$ sensor measurements equally spaced in time, such as accelerations or displacements. Accordingly, the observation space $\mathcal{O}$ is described through the sensitivity of $\mathbf{U}$ with respect to $\boldsymbol{\mu}$. For the problem settings we consider, each observation process refers to a time interval that is short enough for the operational, environmental, and damage conditions to be assumed time-invariant, yet long enough to allow accurate identification of the structural behavior.

As the unknown physical state reflects the ground truth variability in the structural health of the asset, the categorical random variable modeling the digital state describes the presence, location, and magnitude of damage among a set of predefined configurations in $\mathcal{D}$. We describe changes in the structural dynamic response due to damage through a localized reduction of effective stiffness. If the time duration of the observation process is small compared to the time required for damage progression, this approach allows the structural behavior to be treated as linear within an observation interval, thus enabling a timescale separation between damage evolution and health assessment~\cite{Azam_Mariani}. Examples of damage patterns that can be represented in this way include the failure of bolted connections and corrosion in steel structures~\cite{PANDEY19943,tree_ensemble}, cracking and delamination in composite structures~\cite{kapteyn2020data}, cracking in reinforced concrete structures~\cite{tree_ensemble,TEUGHELS2002}, cracking and crushing-induced stiffness losses in masonry brick structures~\cite{Macias}, cuts in steel beams~\cite{seventekidis2020106972}, and the erosion of piles and abrasion of deck surfaces~\cite{fernandeznavamuel2022114016}. Here, the local stiffness reduction shaping $\mathcal{D}$ is achieved by means of two variables, $y\in\mathbb{N}$ and $\delta\in\mathbb{R}$, both included in $\boldsymbol{\mu}$. Specifically, $y\in \mathbb{N}$ labels the specific damage region among a set of $N_\Omega\in\mathbb{N}$ predefined damageable regions, where $y=0$ denotes the undamaged baseline, while $\delta\in\mathbb{R}$ represents the magnitude of the stiffness reduction occurring in the region associated with $y$. 

Data assimilation, represented by the $\phi_t^{\text{data}}$ factor, is performed using DL models to approximate the inverse mapping from measurements $\mathbf{U}$ to the target parameters $y$ and $\delta$. The deterministic output of the DL models is then converted into evidence for configurations in the digital state space that are consistent with the given observation, thereby serving as the likelihood mechanism informing the posterior. This conversion is carried out through a confusion matrix that measures the offline (expected) performance of the DL models in correctly identifying the digital state among the predefined configurations in $\mathcal{D}$. The confusion matrix can be interpreted as a CPT, where rows represent the ground truth digital state, and columns represent the predicted digital state. For a more detailed description of the evaluation of the $\phi_t^{\text{data}}$, the reader is referred to~\cite{Torzoni_DT}.

\section{Online learning of transition dynamics}
\label{sec:methodology}

This describes the proposed adaptive capabilities within the context of parametric MDPs. Specifically, Section~\ref{subsec:online_learning} reviews MDPs to formalize the problem of sequential decision-making under uncertainty and then discusses the online learning of transition dynamics. Section~\ref{subsec:dyn_poli} explores the use of RL to compute control policies based on dynamically inferred transition models. 

\subsection{Parametric Markov decision processes}
\label{subsec:online_learning}

In a decision-making setting, an agent must choose from a set of possible actions, each of which may have uncertain effects on the system state. The goal of the decision-making process is to maximize the expected utilities associated with the outcomes of these actions. Sequential decision problems expand on this concept by requiring the agent to choose a sequence of actions that optimize the long-term accumulation of rewards over the planning horizon.

MDPs provide a framework for modeling sequential decision problems in stochastic environments, using Markov transition models and additive rewards~\cite{Bellman_MDP,dpmp}. A discrete MDP is defined as a $4$-tuple $\langle\mathcal{S},\mathcal{U},\mathcal{P},\mathcal{R}\rangle$, consisting of a set of states $\mathcal{S}$, a set of actions $\mathcal{U}$, a CPT $\mathcal{P}\in\mathbb{R}^{\mathcal{D}\times\mathcal{D}\times\mathcal{U}}$ describing the Markov transition model, and a set of rewards $\mathcal{R}$.

The control action is selected as the best point estimate, thereby mapping the probabilistic state representation to a deterministic decision. This is consistent with the standard MDP assumption of full observability. While alternative approaches account for partial observability (e.g., \cite{papakonstantinou2016point, arcieri_pomdp, arcieri_pomdp2}), they require more complex and computationally demanding solution strategies. In the present work, we rely on the high accuracy of the deep learning models used for digital state classification, which provide a near-complete characterization of the system state. On this basis, we approximate the true state with its most likely estimate, effectively decoupling state estimation from control. We note, however, that this simplification neglects residual uncertainty and may lead to suboptimal decisions when classification uncertainty is non-negligible.

In our DT framework, the MDP associated with optimal control planning is encoded by the prediction DBN shown in Figure~\ref{fig:unrolled_dbn_prediction}. While in~\cite{Torzoni_DT} the transition model $\phi_t^\text{history}$ was kept fixed, we relax the assumption of a stationary Markov process for the conditional dependencies between random variables in consecutive time slices. Specifically, we treat the transition probabilities in $\mathcal{P}$ as random variables with conjugate distributions, similar to~\cite{tezzele2024adaptive}. Starting with an initial prior, the transition dynamics is learned online using Bayesian updates that incorporate new evidence about the system response to actions. This approach is scalable and allows ($i$) adapting DTs starting from (possibly, inaccurate) initial priors or ($ii$) tracking non-stationary, evolving contexts. 

Actions with uncertain effects produce non-deterministic (and possibly unexpected) outcomes. Within a SHM context, inter-state transitions can be described using (stochastic) transition matrices \mbox{$\mathcal{P}_u\in\mathbb{R}^{\mathcal{D}\times\mathcal{D}}$}, one for each action $u\in\mathcal{U}$:
\begin{equation}
\mathcal{P}_u = 
\begin{pmatrix}
p(d_1 \mid  d_1, u) & p(d_1 \mid  d_2, u) & \dots & p(d_1 \mid  d_{|\mathcal{D}|}, u) \\
p(d_2 \mid  d_1, u) & p(d_2 \mid  d_2, u) & \dots & p(d_2 \mid  d_{|\mathcal{D}|}, u) \\
\vdots & \vdots & \ddots & \vdots \\
p(d_{|\mathcal{D}|} \mid  d_1, u) & p(d_{|\mathcal{D}|} \mid  d_2, u) & \dots & p(d_{|\mathcal{D}|} \mid  d_{|\mathcal{D}|}, u)
\end{pmatrix},
\end{equation}
where $|\mathcal{A}|\in\mathbb{N}$ denotes the cardinality of a generic set $\mathcal{A}$. Matrix $\mathcal{P}_u$ collects the probabilities \mbox{$p(d'\mid d,u)$} of transitioning from state $d\in\mathcal{D}$ to state $d'\in\mathcal{D}$, under a given action $u\in\mathcal{U}$. Here, columns represent the starting state $d$, and rows represent the arrival state $d'$. Each column of $\mathcal{P}_u$ sums to one, reflecting the exhaustive set of transition probabilities from a given state $d$ under the action $u$. By partitioning $\mathcal{D}$ in lexicographic order, with states ordered first by damage location and then by damage level, the diagonal entries in $\mathcal{P}_u$ denote the probability the system has of remaining in the same state, while the lower-left and upper-right triangles represent the probabilities of system deterioration and improvement, respectively. The CPT describing the complete transition model, tensorial with respect to all available actions, is given by $\mathcal{P}=\lbrace\mathcal{P}_{1}\ldots,\mathcal{P}_{|\mathcal{U}|}\rbrace$.

The specific content of the transition matrices should account for the structural system under consideration, its operational conditions, and the surrounding environmental variability~\cite{temperatura}. To address these sources of uncertainty, we adopt a hierarchical Dirichlet-Multinomial parametrization. For each state–action pair, the categorical transition distribution is endowed with a Dirichlet prior, whose hyperparameters encode prior beliefs over transition outcomes. This defines a hierarchical Bayesian structure in which transition probabilities are treated as latent random variables, and uncertainty is propagated through the higher-level Dirichlet prior, capturing both epistemic uncertainty and data-driven adaptation~\cite{gelman2013bayesian}. These parameters are then subject to temporal updating, i.e., learned online as new data arrives, enabling adaptive Bayesian inference. 

We model the transition probabilities $p(d'\mid d,u)$ with categorical distributions parameterized by
$\boldsymbol{\theta}_{d,u} = (\theta_{d,u}^{1}, \dots, \theta_{d,u}^{|\mathcal{D}|})^\top\in\mathbb{R}^{|\mathcal{D}|}$, where $\sum_{k=1}^{|\mathcal{D}|}\theta_{d,u}^{k} = 1$ and $k=1,\ldots,|\mathcal{D}|$ indexes the arrival states $d'$. The uncertainty over $\boldsymbol{\theta}_{d,u}$ is represented by a Dirichlet prior:
\begin{equation}
\boldsymbol{\theta}_{d,u}\sim\text{Dirichlet}(\boldsymbol{\alpha}_{d,u}),
\label{eq:dir}
\end{equation}
where $\boldsymbol{\alpha}_{d,u} = (\alpha_{d,u}^{1}, \dots, \alpha_{d,u}^{|\mathcal{D}|})^\top\in\mathbb{R}^{|\mathcal{D}|}$ are (positive) concentration parameters encoding prior beliefs about the transition probabilities. We use the Dirichlet distribution because it is conjugate to the Multinomial likelihood, ensuring that the posterior distribution remains within the Dirichlet family after incorporating observations. This property allows efficient refinement of the transition probabilities beliefs by updating the Dirichlet hyperpriors in closed form.

Managing $|\mathcal{D}| \times |\mathcal{U}|$ Dirichlet distributions, each with $|\mathcal{D}|$ parameters, can become computationally intractable for large state spaces. While we provide the formulation and update scheme for such a state-dependent transition model in \ref{sec:state_dep}, a more computationally viable approach to handling multi-step transitions is to assign equal transition probabilities between consecutive elements of the partition, rather than a separate prior for each individual state transition~\cite{andriotis2019managing,ross2011bayesianadaptive,papakonstantinou2016point, tezzele2024adaptive}. In other words, we assign probabilities of remaining in the same state, of transitioning by one step, two steps, and so on, independently of the starting state $d$. This state-independent transition model is thus parametrized through \mbox{$\mathbf{p}_u = (p_u^0,p_u^1,\ldots,p_u^I)^\top\in\mathbb{R}^{I+1}$}, representing transitions probabilities of $0,1,\ldots,I$ steps forward under action $u$. The corresponding transition matrix $\mathcal{P}_u$ is then constructed from $\mathbf{p}_u$, ensuring the same column structure for all starting states, except for absorbing states --- i.e., states from which no further transitions occur, meaning they retain probability $1$ of remaining in the same state.

As an example, for a state space of cardinality $|\mathcal{D}| = 5$ with a single damage location, the transition matrix for an action under which the structural health state can deteriorate by up to $I=2$ steps is:
\begin{equation}
\mathcal{P}_u(\mathbf{p}_u) = 
\begin{pmatrix}
p_u^0 & 0 & 0 & 0 & 0\\
p_u^1 & p_u^0 & 0 & 0 & 0\\
p_u^2 & p_u^1 & p_u^0 & 0 & 0\\
0 & p_u^2 & p_u^1 & p_u^0 & 0\\
0 & 0 & p_u^2 & p_u^1 + p_u^2 & 1\\
\end{pmatrix}.
\end{equation}
As per Equation~\eqref{eq:dir}, the probabilities $\mathbf{p}_u$ are assigned a Dirichlet prior:
\begin{equation}
\mathbf{p}_u \sim \text{Dirichlet}(\boldsymbol{\alpha}_u),
\end{equation}
where $\boldsymbol{\alpha}_u = (\alpha_u^0, \alpha_u^1, \ldots, \alpha_u^I)^\top\in\mathbb{R}^{I+1}$ are the concentration parameters for the step transitions under action $u$.

Let $d_{0:t_c}$ and $u_{0:t_c-1}$ denote the history of system states and actions observed up to the current time $t_c$, respectively. Additionally, let $\mathbf{y}_u=(N^{0}_{u},N^{1}_{u},\ldots,N^{I}_{u})^\top\in\mathbb{N}^{I+1}$ represent the counts of observed step transitions under $\lbrace d_{0:t_c},u_{0:t_c-1}\rbrace$. The likelihood of $\mathbf{y}_u$, given the transition probabilities $\mathbf{p}_u$, is modeled as a Multinomial distribution:  
\begin{equation}
\mathbf{y}_u\mid  \mathbf{p}_u \sim \text{Multinomial}(N_u, \mathbf{p}_u),
\end{equation}
where $N_u = \sum_{i=0}^I N_u^i$ is the total transitions count observed under action $u$. Using Bayes' theorem, the posterior distribution of $\mathbf{p}_u$ after observing $\mathbf{y}_u$ remains in the Dirichlet family due to conjugacy:
\begin{equation}
\mathbf{p}_u\mid \mathbf{y}_u \sim \text{Dirichlet}(\overline{\boldsymbol{\alpha}}_{u}),
\end{equation}
with updated concentration parameters $\overline{\boldsymbol{\alpha}}_{u}\in\mathbb{R}^{I+1}$, computed as:
\begin{equation}
\overline{\boldsymbol{\alpha}}_{u} = (\alpha_u^0 + N_u^0, \alpha_u^1 + N_u^1, \ldots, \alpha_u^I+ N_u^I)^\top.
\end{equation}

The parametrized Dirichlet-Multinomial transition model efficiently accounts for the possibility of multi-step transitions. However, it does not address the possibility of inconsistencies arising from evidence that contradicts the transition model. This limitation can be particularly critical, as assigning a zero probability to potential state transitions could compromise the inference process~\cite{lindley2013understanding}. To resolve this potential inconsistency, a perturbation matrix $\Xi\in\mathbb{R}^{|\mathcal{D}|\times|\mathcal{D}|}$, with positive entries, arbitrary small relative to the non-zero terms of $\mathcal{P}_u$, is introduced. This matrix is used to obtain a perturbed transition matrix as follows: 
\begin{equation}
    \mathcal{P}_u^\Xi(\mathbf{p}_u) = \mathcal{P}_u(\mathbf{p}_u) + \Xi ,
    \label{eq:perturbed}
\end{equation}
which is then properly normalized. 

A special case of the state-independent Dirichlet-Multinomial transition model is the Beta-Binomial model, which is parameterized by a single 1-step transition probability. The formal derivation follows an approach analogous to the one presented in this section and is provided in \ref{sec:beta_bern} for completeness. The numerical experiments in Section~\ref{sec:res} use Dirichlet-Multinomial state-independent transition models.

\subsection{Dynamic control policies and precision updates} \label{subsec:dyn_poli}

The $\phi_t^{\text{control}}$ factor in Equation~\eqref{eq:joint_factor} represents a CPT encoding a health-aware control policy $\pi_t(D_t)$. In~\cite{Torzoni_DT}, a stationary policy was computed offline for an infinite planning horizon, based on the simplifying assumption of known transition dynamics. Here, the sequence of actions $u^*_{t_c:T}$ defining the optimal control policy $\pi^*$ is systematically recomputed using MBRL with a dynamically updated transition model. The policy is recomputed over the finite planning horizon $[t_c,T]$, representing the design lifetime of the asset. The corresponding optimization problem is expressed as:
\begin{equation}
    \pi_t^*(D_t) = \argmax_{\pi_t}\,{\mathbb{E}\Bigg[\sum_{t=t_c}^{T}{R_{t}}\Bigg]}.
\label{eq:finite_hor_optimal}
\end{equation}
The subscript $t$ in $\pi_t^*$ denotes a non-stationary control policy, as the optimal action depends on the remaining time before the end of the planning horizon, as well as the time $t_c$ during which the transition dynamics is updated and the control policy is recomputed. The additive reward function to be maximized is defined as follows:
\begin{equation}
    R(d,u) = R^\text{health}(d) + \xi R^\text{control}(u), \qquad \forall d \in \mathcal{D}, \, \forall u \in \mathcal{U},
    \label{eq:reward}
\end{equation}
where $R^\text{health}(d)$ and $R^\text{control}(u)$ quantify the rewards associated with the health state of the system and the action taken, respectively. In general, these reward values may represent real-world costs or abstract values, tuned using the weighting factor $\xi\in\mathbb{R}$ to achieve the desired system performance.

One way to characterize an MDP is to consider the expected utility associated with a policy $\pi$ when starting in a certain state and following $\pi$ thereafter. For a finite horizon $T$, the state-value function $V^\pi:\mathcal{D}\to\mathbb{R}$ quantifies, for every state and starting time $t$, the expected cumulative reward when starting at $d_{t}$ and following policy $\pi$, as follows~\cite{Sutton}:
\begin{equation}
V^\pi(d_{t}) = \mathbb{E}_\pi\Bigg[\sum_{\tau=t}^{T}{R_{\tau}}\,\,\Bigg|\,\,d_{t}\Bigg],\label{eq:state-value}
\end{equation}
where $\mathbb{E}_\pi[\,\cdot\,]$ denotes the expected value, given that the agent follows policy $\pi$. For any control policy $\pi$ and for each possible state $d_t$, with $t<T$, the state-value function satisfies the following recursive relationship, known as the Bellman equation for $V^\pi$:
\begin{equation}
V^\pi(d_t) = \sum_{u\in\mathcal{U}}\pi(u\mid d_t)\sum_{d_{t+1}\in\mathcal{D}}p(d_{t+1}\mid d_t,u)\bigl(r_t+V^\pi(d_{t+1})\bigr),
\label{eq:consistenz}
\end{equation}
where $r_t$ is the immediate reward at time $t$, and $p(d_{t+1}\mid d_t,u_t)\in\mathcal{P}$ are the transition probabilities. According to the principle of maximum expected utility, for each time $t<T$, there exists a time-dependent optimal policy $\pi_t^*$ that maximizes expected utility. The time-dependent optimal action at state $d_t$ is given as:
\begin{equation}
\pi_t^*(d_t) = \argmax_{u\in\mathcal{U}}\Bigg(\sum_{d_{t+1}\in\mathcal{D}}p(d_{t+1}\mid d_t,u)\bigl(r_t+V^{*}(d_{t+1})\bigr)\Bigg),
\label{eq:optimiz_2}
\end{equation}
where $V^{*}(d_{t+1})$ measures the utility associated with $d_{t+1}$ when following $\pi^*$. The Bellman optimality equation for $\pi^*$ can thus be written as:  
\begin{equation}
V^{\pi*}(d_t) = \max_{u\in\mathcal{U}}\Bigg(\sum_{d_{t+1}\in\mathcal{D}}p(d_{t+1}\mid d_t,u)\bigl(r_t+V^{*}(d_{t+1})\bigr)\Bigg),
\label{eq:Bellman}
\end{equation}
yielding a system of $|\mathcal{D}|$ recursions in $|\mathcal{D}|$ unknowns. The optimal state-value functions are the unique solutions to these equations, from which the corresponding optimal policy can be directly derived. However, the optimal action $u_t^*$ is selected through the $\max$ operator, which introduces nonlinearity in the system of Bellman equations. 

An approximation of the unique solution to Equation~\eqref{eq:Bellman} is computed using finite-horizon dynamic programming. Dynamic programming is an algorithmic paradigm introduced by Bellman~\cite{bellman1957} that simplifies the optimization by decomposing it into smaller sub-problems that are solved recursively. Among these methods, we employ the value iteration algorithm, also known as backward induction, as it proceeds by solving the problem backward in time, from $t=T$ down to~$t=t_c$.

An algorithmic description of the offline and online phases of the proposed computational procedure are reported in Algorithm~\ref{alg:offline} and Algorithm~\ref{alg:finite_MBRL}, respectively.
 
\begin{algorithm}[ht]
\hspace*{\algorithmicindent} \textbf{Input}: parametrization of the physics-based models\\
\hspace*{44pt} training options for the reduced-order model\\
\hspace*{44pt} training options for the deep learning models
\begin{algorithmic}[1]
\State set up the physics-based numerical model of the structure
\State assemble snapshot matrix of full-order solutions
\State find reduced basis functions via singular value decomposition of the snapshots matrix
\State use the reduced-order model to populate the training dataset with recordings at sensor locations
\State train and validate the deep learning models for structural health estimation
\State compute the confusion matrix associated with $\phi_t^{\text{data}}$
\State set the prior distributions over the transition probabilities  defining $\phi_t^\text{history}$
\State set the reward function encoding $\phi_t^\text{reward}$
\end{algorithmic}
\hspace*{\algorithmicindent} \Return trained deep learning models for state estimation\\
\hspace*{45.4pt} topology and factors for the online phase DBN\\
\hspace*{45.4pt} topology and factors for the prediction DBN
\caption{Preliminary offline phase.}
\label{alg:offline}
\end{algorithm}
\begin{algorithm}[ht]
\hspace*{\algorithmicindent} \textbf{input}: 
topology and factors for the online phase DBN\\
\hspace*{43pt} topology and factors for the prediction DBN\\
\hspace*{43pt} trained deep learning models for state estimation\\
\hspace*{43pt} observational data $O_{t_c}=o_{t_c}$\\
\hspace*{43pt} history of observed system states and actions $\lbrace d_{0:t_c},u_{0:t_c-1}\rbrace$
\begin{algorithmic}[1]
\State $t_c\leftarrow t_c+1$
\State assimilate $o_{t_c}$ with deep learning models
\State infer $D_{t_c}$, given $D_{t_c-1}$ and $u_{t_c-1}$
\State adapt transition matrices $\mathcal{P}_u(\mathbf{p}_u)$ by updating $p(\mathbf{p}_u\mid \lbrace d_{0:t_c},u_{0:t_c-1}\rbrace)$
\State \textbf{optional}: choose the statistic parametrizing the transition model to adjust risk aversion
\State recompute control policy $\pi = (\pi_{t_c},\ldots, \pi_{T})$ under the updated transition model
\State forecast $D_{t_c+1:t_p}$ and $U_{t_c+1:t_p}$ using the updated transition model and control policy
\State select $u_{t_c}$ as the best point estimate of $U_{t_c}$
\end{algorithmic}
\hspace*{\algorithmicindent} \Return control action to be executed $U_{t_c}=u_{t_c}$\\
\hspace*{45.4pt} expected evolution $D_{t_c+1:t_p}$\\
\hspace*{45.4pt} expected evolution $U_{t_c+1:t_p}$\\
\hspace*{45.4pt} updated transition model CPT $\mathcal{P}$\\ 
\hspace*{45.4pt} updated control policy $\pi\approx\pi^*$
\caption{Online phase.}
\label{alg:finite_MBRL}
\end{algorithm} 

\paragraph{Remark}%
For each action, the transition matrix $\mathcal{P}_u$ is parametrized by a statistic of the inferred transition posteriors leading to different control policies. The parametrizing statistic is directly connected to the risk aversion of the decision-maker which involves balancing a trade-off between maximizing profits and ensuring structural safety. In the numerical experiments presented in Section~\ref{sec:res}, the transition matrices are parameterized using the maximum a-posteriori estimate of the inferred Dirichlet posterior. Alternative choices could be adopted to induce different levels of risk aversion. For univariate probability distributions it is possible to incorporate different risk measures as proposed in~\cite{tezzele2024adaptive}. For the Dirichlet distribution, due to its multivariate nature and its support on simplexes, it is not possible to extend commonly used risk measures such as the value at risk~\cite{morgan} or the conditional value at risk~\cite{cvar_cite} without using an intermediate scalar function.

\section{Numerical experiments}\label{sec:res}

This section assesses the proposed methodology on the simulated monitoring, management, and maintenance planning of the H{\"o}rnefors railway bridge.

\subsection{Physical asset}

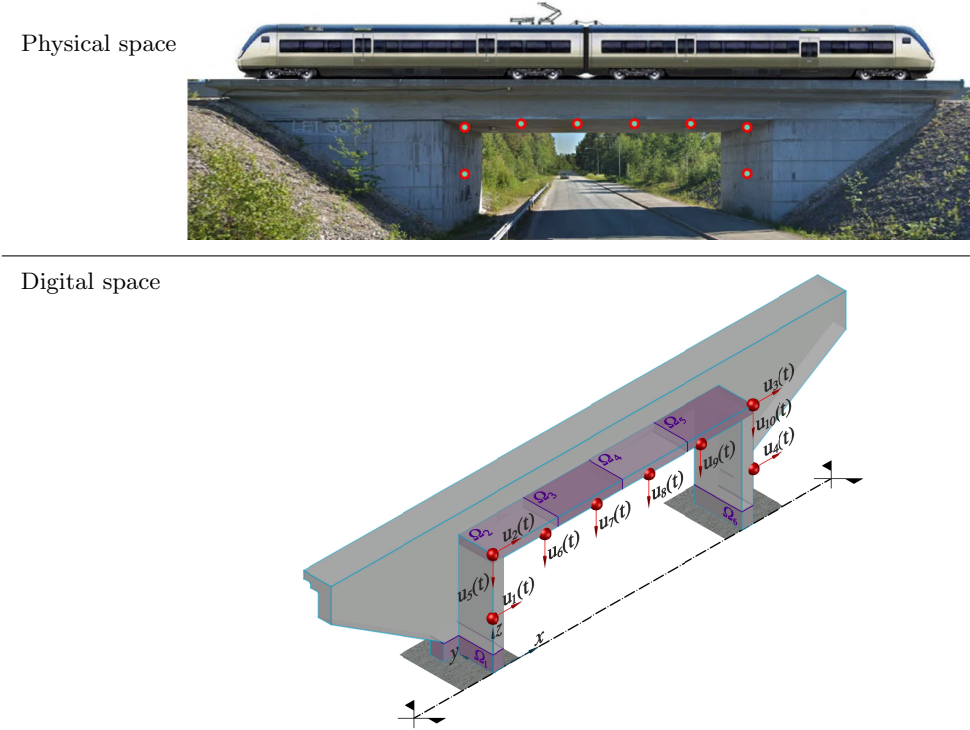
\begin{figure}[ht]
\center
\begin{tikzpicture}[scale=.9, every node/.style={scale=1.}]

\node[draw=none,fill=none] at (6,1.9){\includegraphics[trim=0 25 0 0, clip, width=0.7\textwidth]{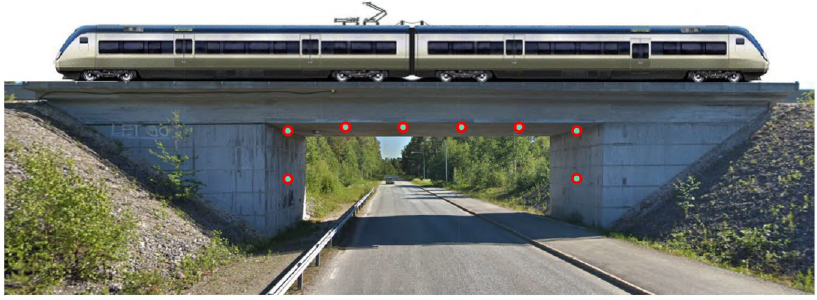}};

\node[draw=none,fill=none] at (6,-3.7){\includegraphics[width=0.5\textwidth, angle=0]{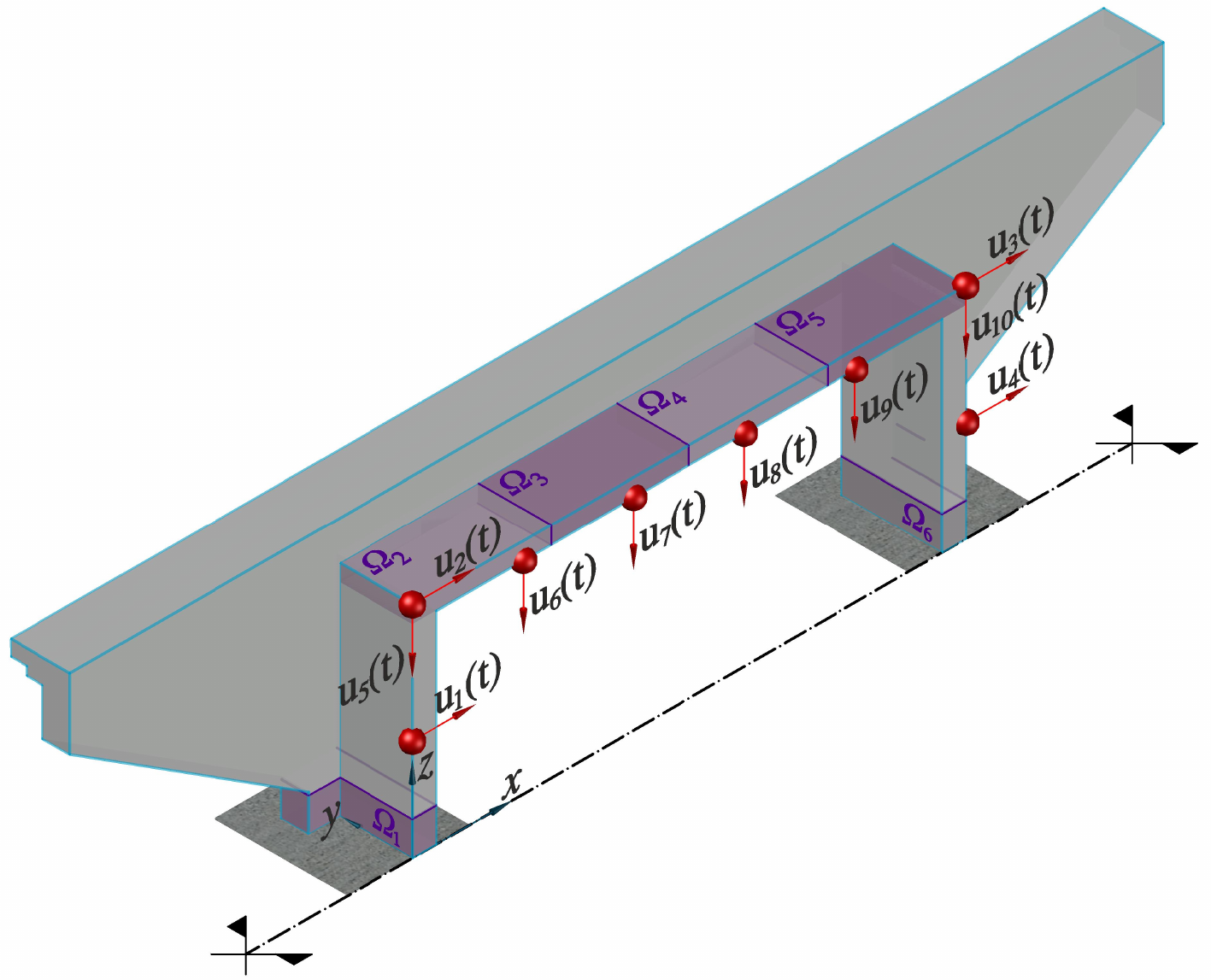}};

\draw[black] (-2.5,-0.1) to (13.5,-0.1);
\node [] () at (-1.08,3) {\small Physical space};
\node [] () at (-1.2,-0.5) {\small Digital space};

\end{tikzpicture}

\caption{Physical asset and its digital twin - The physical space corresponds to the H{\"o}rnefors railway bridge. The digital space is a schematization for structural health monitoring, including details of synthetic recordings related to displacements $u_1(t),\ldots,u_{10}(t)$, and predefined damage regions $\Omega_1,\ldots,\Omega_6$.} 
\label{fig:bridge_plus_digital}
\end{figure}

The H{\"o}rnefors railway bridge, shown in Figure~\ref{fig:bridge_plus_digital},
is an integral concrete bridge located along the Swedish Bothnia line. The bridge features a span of $15.7~\text{m}$, a free height of $4.7~\text{m}$, and a width of $5.9~\text{m}$ (edge beams excluded). The structural elements have a thicknesses of $0.5~\text{m}$ for the deck, $0.7~\text{m}$ for the frame walls, and $0.8~\text{m}$ for the wing walls. The bridge is founded on two plates, connected by two stay beams and supported by pile groups. The concrete is of class C35/45, with mechanical properties: $E=34~\text{GPa}$, $\nu= 0.2$, $\rho=2500~\text{kg/m}^3$. The superstructure consists of a single track with sleepers spaced $0.65~\text{m}$ apart, resting on a $0.6~\text{m}$ deep, $4.3~\text{m}$ wide ballast layer with a density of $\rho_B=1800~\text{kg/m}^3$. The bridge is subjected to the transit of \textit{Gr{\"o}na T{\r a}get} trains traveling at speeds ranging from $v\in[160,215]~\text{km/h}$. We specifically consider trains composed of two wagons, totaling $8$ axles, with each axle carrying a mass of $\psi\in[16,22]~\text{ton}$. The geometrical and mechanical data, as well as the traveling load model, are adapted from~\cite{Torzoni_DT,metodologico}.

The integral design of the H{\"o}rnefors bridge minimizes the ingress of moisture, debris, and corrosive agents, thereby enhancing its long-term durability. The types of damage that can affect such bridges depend on material and structural factors, environmental conditions, and traffic loads. Some damage patterns that can be described through a localized stiffness reduction include: concrete cracking caused by thermal gradients, freeze-thaw cycles, or overloading; gradual concrete degradation due to the chemical reaction between reactive silica aggregates and alkalis in the cement paste, potentially resulting in cracking and spalling; concrete cracking and structural damage induced by stress concentrations from differential foundational settlements; erosion of concrete surfaces due to vehicular collisions or prolonged exposure to environmental factors. 

\subsection{Dataset assembly and training options}

\begin{figure}[ht]
\centering
  \includegraphics[width=.9\textwidth]{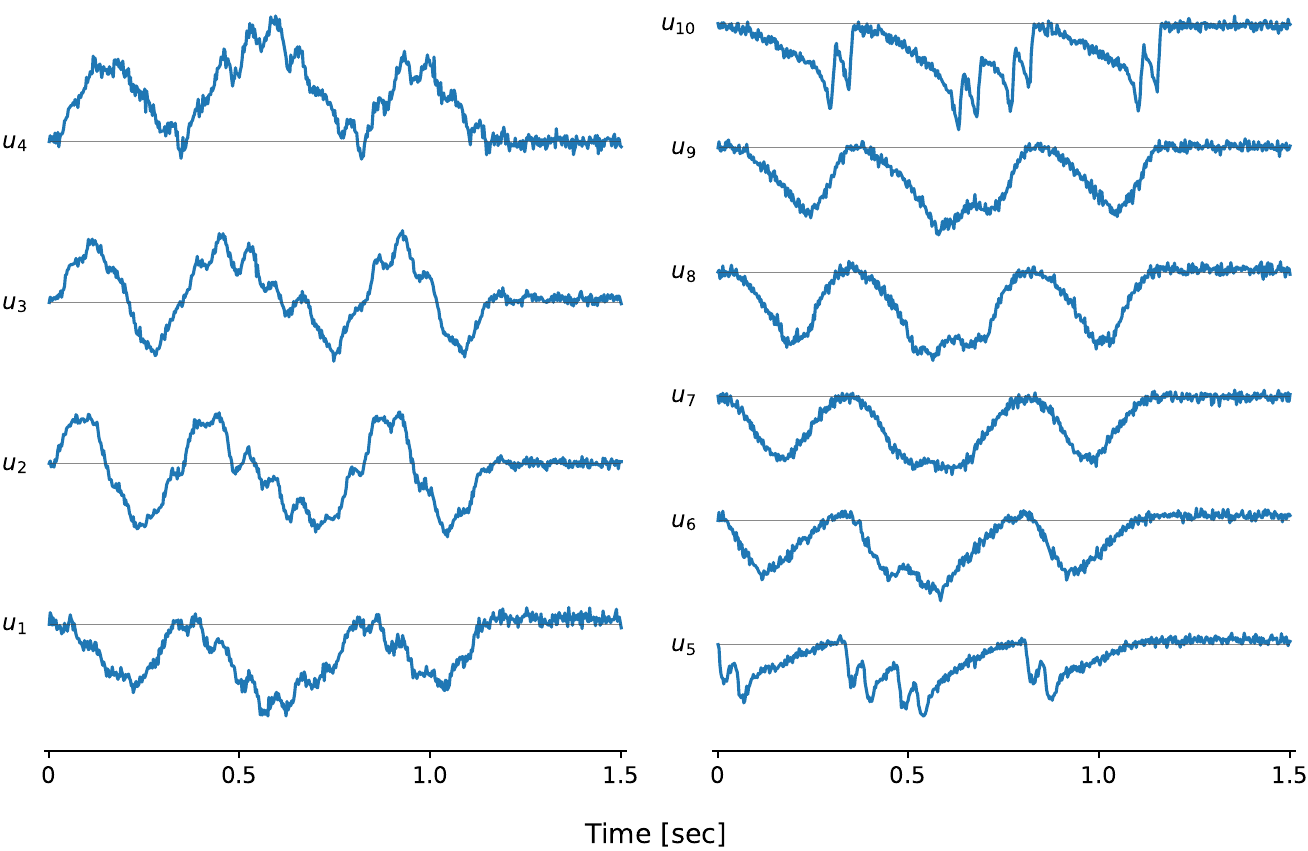}
  \caption{Exemplary multivariate time series of structural vibrations (standardized displacements $u_1(t),\ldots,u_{10}(t)$) induced by the transit of a pilot train across the railway bridge.}
  \label{fig:vib_ts}
\end{figure}

The monitoring system is deployed to provide displacement recordings $\mathbf{U}$, corresponding to the $N_s=10$ degrees of freedom highlighted in Figure~\ref{fig:bridge_plus_digital}. Each recording spans a time interval $[0,T=1.5~\text{s}]$, with a sampling frequency of $400~\text{Hz}$. To simulate measurement errors caused by sensor self-noise, the recordings are corrupted with an additive Gaussian noise to yield a signal-to-noise ratio of $120$. An exemplary set of displacement time histories is shown in Figure~\ref{fig:vib_ts}.

The parameter $y\in\lbrace0,\ldots,6\rbrace$ labels the specific damage region. In addition to the damage-free baseline $y=0$, the presence of damage in the structure is simulated by considering a set of $N_\Omega=6$ predefined damageable subdomains $\Omega_m$, with $m=1,\ldots,6$, as depicted in Figure~\ref{fig:bridge_plus_digital} and corresponding to $y=1,\ldots,6$. Within these regions, a localized reduction of the material stiffness can occur with a magnitude $\delta\in[30\%,80\%]$, and is kept fixed while a train travels across the bridge. 

The full-order model is the same as that described in~\cite{Torzoni_DT}. It features $17,292$ degrees of freedom, resulting from a space discretization using linear tetrahedral finite elements with an element size of $0.80~\textup{m}$, reduced to $0.15~\textup{m}$ for the deck. The ballast layer is modeled through an increased density for the deck and the edge beams. The embankments are modeled using distributed springs over the surfaces in contact with the ground, implemented as a Robin mixed boundary condition (with elastic coefficient $10^{8}~\textup{N/m}^3$). The structural dissipation is modeled using a Rayleigh's damping matrix, calibrated to achieve a $5\%$ damping ratio for the first two structural modes. 

The solution time interval is uniformly partitioned into $600$ time steps, and the time integration is performed using an implicit Newmark scheme~\cite{hughes2000finite}. We use the (unconditionally stable) constant average acceleration rule, so that the integration step can be arbitrarily selected based on the structural frequencies and the sensor sampling rate.

To reduce the computational cost of solving the full-order model for any sampled $\boldsymbol{\mu}$, a projection-based reduced-order model is used to speed up the dataset generation for training the DL models. We use a proper orthogonal decomposition Galerkin reduced basis method~\cite{RB, chinestaenc2017}. The reduced-order model is derived from a snapshot matrix assembled from $400$ full-order solutions for varying input parameters $\boldsymbol{\mu} = (v,\psi,y,\delta)^\top$, taken as uniformly distributed and sampled via Latin hypercube rule. The order of the reduced-order expansion is determined using an energy criterion. By prescribing a tolerance of $10^{-3}$ for the fraction of energy content to be disregarded, the number of degrees of freedom is reduced to $133$. Using this reduced-order model, the training dataset is populated with $10,000$ instances. These data are subsequently used to train DL models for detecting, locating, and quantifying structural damage, as described in~\cite{Torzoni_DT}.

The application of DL in SHM follows a pattern recognition paradigm~\cite{Bishop}, where damage is assessed by comparing incoming measurements with data previously collected under known structural conditions. By capturing temporal correlations within and across time series data, DL optimizes the extraction of damage-sensitive features for modeling the relationship between inputs and damage states. Moreover, as DL models are trained offline, they provide real-time predictions during operation.

Within our framework, each time new observational data are acquired --- when a train crosses the bridge, marking a new DT time slice --- they are first processed by a classification model to address damage detection and localization. The target class corresponds to one of the predefined damage scenarios described by the $y$ parameter. The classifier maps the vibration recordings $\mathbf{U}$ onto confidence levels representing the likelihood of $\mathbf{U}$ belonging to each damage class. The most likely class is selected as the best point estimate for categorizing the measurements. Such classifier is trained by minimizing the categorical cross-entropy loss function between the one-hot encoding of $y$ and its predicted counterpart.

Whenever damage is detected and localized in the $m$-th region $\Omega_m$, where $m=1,\ldots,6$, the observational data are further processed with a regression model --- one for each damageable region --- to quantify the extent of damage $\delta$. Each regression model maps the vibration recordings associated with the $m$-th damage region onto the estimated magnitude of the stiffness reduction within that region. These regressors are trained by minimizing a mean squared error loss function between the target stiffness reduction and its predicted value. For a detailed and formal description of the DL models architectures and their implementation, the reader is referred to~\cite{Torzoni_DT}.

\subsection{Digital twin framework}

Since the space of digital states is discrete, the outcomes of the DL models are accommodated into the PGM by discretizing the value range of $\delta$ into $N_\delta\in\mathbb{N}$ intervals. The number and width of these intervals are chosen arbitrarily, without specific restrictions. The $\delta$ range is discretized int o $N_\delta=6$ intervals $\lbrace[30\%,35\%],$ $[35\%,45\%],$ $[45\%,55\%],$ $[55\%,65\%],$ $[65\%,75\%],$ $[75\%,80\%]\rbrace$. This results in $|\mathcal{D}|=1+N_\delta N_\Omega=37$ digital states, ordered first by damage location and then by damage level.

In the absence of experimental data, the testing phase of the DL models is carried out using $4000$ noisy full-order solutions. The ground truth digital state is classified with an accuracy of $91.39\%$.

We consider four different control actions, each associated with a CPT that models the transition probability $p(D_{t+1} \mid  D_t, U_t = u_t)$. Together, these CPTs encode the $\phi^\text{history}_t$ factor. Note that these internal models predict the evolution of structural health but do not replicate the ground truth progression, which remains unknown to the DT. The considered control actions are as follows:
\begin{enumerate}
    \item \textit{Do nothing} (DN) action: no maintenance is scheduled, and the structural health state evolves according to a stochastic deterioration process, while regular revenue is maintained.    
    \item \textit{Restrict operations} (RO) action: traffic flow on the bridge is restricted to lightweight trains carrying less than $18~\text{ton}$ per axle. This reduces the structure degradation rate but also decreases the revenue generated by the infrastructure.
    \item \textit{Minor repair} (MR) action: a moderate-cost maintenance intervention is carried out to address specific damages, though the structure does not return to a damage-free state.
    \item \textit{Perfect repair} (PR) action: a high-cost maintenance intervention is performed, almost certainly restoring the bridge to a damage-free state.    
\end{enumerate}

The (unknown) ground truth stochastic model varies depending on the most recent control actions. We assume that damage can develop in any of the predefined regions, without propagating between different damageable subdomains. Based on this assumption, we define below the prescribed degradation (or improvement) evolution model associated with each action.

\paragraph{Do nothing action} In the absence of maintenance actions, a (simulated) stochastic deterioration process monotonically degrades the structural health. If the (unknown) state of the system is undamaged, the presence/location of damage $y$ is sampled from a categorical distribution. This distribution assigns half of its probability mass to the undamaged configuration $y=0$, indicating that the bridge remains intact. The remaining half is evenly distributed among the remaining damage classes $y\neq0$. Damage may develop with magnitude $\delta$ uniformly sampled within the range of the first damage interval $[30\%,35\%]$. In contrast, if the system is already damaged, further deterioration propagates through $\delta$ increments sampled from a categorical distribution with Dirichlet hyperparameters $\boldsymbol{\alpha}_\text{DN}^\text{true}=(\alpha_\text{DN}^\text{0,true}, \alpha_\text{DN}^\text{1,true}, \alpha_\text{DN}^\text{2,true})^\top$, scaled by the $\delta$ discretization step $10\%$, and finally perturbed by Gaussian noise with mean $2\%$ and standard deviation $0.5\%$ (negative increments are rounded to zero). The DN model can be formalized as follows:
\begin{align}
    y &\sim \text{Categorical}\left (\frac{1}{2}, \frac{1}{12}, \frac{1}{12}, \frac{1}{12}, \frac{1}{12}, \frac{1}{12}, \frac{1}{12}\right ),\\
    \delta_0 \mid  y\neq0 &\sim \text{Uniform}(0.3, 0.35),\\
    \delta_t - \delta_{t-1} \mid  y\neq0&\sim 0.1\cdot\text{Categorical}(\mathbf{p}_\text{DN}^\text{true})+\text{Normal}_{\geq 0}(0.02, 0.005),\\\mathbf{p}_\text{DN}^\text{true}&\sim\text{Dirichlet}(\boldsymbol{\alpha}_\text{DN}^\text{true}),\quad \boldsymbol{\alpha}_\text{DN}^\text{true}=(500, 400, 200)^\top.
\end{align}
The prescribed trajectory of the structural health parameters is arbitrarily chosen to fully showcase the capabilities of the DT. However, the DT would be equally capable of tracking the structural health evolution, whether considering more or less aggressive degradation processes.

\paragraph{Restrict operation action} If only lightweight trains are allowed to cross the bridge, the model resembles the one for the DN action but is characterized by a lower probability of damage initiation and a reduced deterioration rate. In this scenario, damage may occur with a probability of $0.25$, evenly distributed among the damage classes. Once damage develops, it propagates with $\delta$ increments sampled from a categorical distribution with Dirichlet hyperparameters $\boldsymbol{\alpha}_\text{RO}^\text{true}=(\alpha_\text{RO}^\text{0,true}, \alpha_\text{RO}^\text{1,true}, \alpha_\text{RO}^\text{2,true})^\top$, scaled by the $\delta$ discretization step $10\%$, and finally perturbed by Gaussian noise with mean $2\%$ and standard deviation $0.5\%$ (negative increments are rounded to zero). The RO model is formalized as follows:
\begin{align}
    y &\sim \text{Categorical}\left(\frac{3}{4}, \frac{1}{24}, \frac{1}{24}, \frac{1}{24}, \frac{1}{24}, \frac{1}{24}, \frac{1}{24}\right),\\
    \delta_0 \mid  y\neq0 & \sim \text{Uniform}(0.3, 0.35),\\
      \delta_t - \delta_{t-1} \mid  y\neq0&\sim 0.1\cdot\text{Categorical}(\mathbf{p}_\text{RO}^\text{true})+\text{Normal}_{\geq 0}(0.02, 0.005),\\\mathbf{p}_\text{RO}^\text{true}&\sim\text{Dirichlet}(\boldsymbol{\alpha}_\text{RO}^\text{true}),\quad \boldsymbol{\alpha}_\text{RO}^\text{true}=(600, 300, 100)^\top.
\end{align}

\paragraph{Minor repair action} If the structural health state of the bridge undergoes partial improvement and $y=0$, the system remains undamaged. If $y\neq0$, the damage magnitude $\delta$ decreases. Specifically, the $\delta$ decrement is sampled from a categorical distribution with Dirichlet hyperparameters $\boldsymbol{\alpha}_\text{MR}^\text{true}=(\alpha_\text{MR}^\text{0,true}, \alpha_\text{MR}^\text{0,true}, \alpha_\text{MR}^\text{0,true})^\top$, scaled by the $\delta$ discretization step $10\%$, and finally perturbed by Gaussian noise with mean $-2\%$ and standard deviation $0.5\%$ (positive increments are rounded to zero). The state is updated to reflect a partial repair, ensuring that the damage magnitude does not fall below the undamaged state threshold of $\delta=30\%$. The MR model is formalized as follows:
\begin{equation}
    \begin{aligned}
        \delta_t - \delta_{t-1} \mid  y\neq0&\sim -0.1\cdot\text{Categorical}(\mathbf{p}_\text{MR}^\text{true})+\text{Normal}_{\leq 0}(-0.02, 0.005),\\\mathbf{p}_\text{MR}^\text{true}&\sim\text{Dirichlet}(\boldsymbol{\alpha}_\text{MR}^\text{true}),\quad \boldsymbol{\alpha}_\text{MR}^\text{true}=(1, 100, 300)^\top.
    \end{aligned}
\end{equation}

\paragraph{Perfect repair action} Under PR, the system returns to the damage-free configuration, independently of its damage state.

\paragraph{Transition model}
All numerical results are obtained using a Dirichlet–Multinomial state-independent transition model. Transition probabilities thus depend only on the selected action and not on the current state, and are learned online through action-specific Dirichlet priors updated via observed state transitions.

The assumed transition model $p(D_{t+1} \mid  D_t, U_t = u_t)$ satisfies the assumption that only one damage scenario can be activated at a time. The transition dynamics associated with the different actions are formulated as described in Section~\ref{subsec:online_learning}. Starting from an initial belief over the transition probabilities, these are subsequently refined by integrating evidence about the system response to applied actions. Specifically, we instantiate two Dirichlet-Multinomial state-independent models to represent degradations of zero, one, or two $\delta$ intervals under the DN and RO actions, respectively, and one Dirichlet-Multinomial state-independent model to represent improvements of zero, one, or two $\delta$ intervals under the MR action. The initial Dirichlet hyperpriors are assigned as:
\begin{align}
    \boldsymbol{\alpha}_\text{DN}=&(\alpha_\text{DN}^0, \alpha_\text{DN}^1, \alpha_\text{DN}^2)^\top=(10, 7, 4)^\top,\\
        \boldsymbol{\alpha}_\text{RO}=&(\alpha_\text{RO}^0, \alpha_\text{RO}^1, \alpha_\text{RO}^2)^\top=(10, 5, 2)^\top,\\
            \boldsymbol{\alpha}_\text{MR}=&(\alpha_\text{MR}^0, \alpha_\text{MR}^1, \alpha_\text{MR}^2)^\top=(3, 10, 9)^\top,
\end{align}
where $\boldsymbol{\alpha}_\text{DN}$ and $\boldsymbol{\alpha}_\text{RO}$ prioritize remaining in the same state over 1-step and 2-step degrading transitions, while $\boldsymbol{\alpha}_\text{MR}$ prioritizes of 1-step and 2-step recovery over remaining in the same state. Finally, for the PM action, we prescribe a transition matrix that assigns unit probability to transitioning to the undamaged state, independently of the current
condition.

Transitions are defined in terms of discrete step variations, corresponding to changes of zero, one, or two $\delta$ intervals. The state space is bounded by construction, so that any transition exceeding the admissible range is mapped onto the nearest valid state, i.e., the undamaged or most-damaged condition. Moreover, it is worth highlighting that our state-independent assumption is appropriate when the deterioration and repair dynamics are primarily governed by the applied action rather than the current damage level, or when available data are insufficient to reliably estimate fully state-dependent transitions. Under these conditions, pooling observations across states improves statistical efficiency and yields more stable learning, while still capturing the dominant degradation and recovery trends.

\paragraph{Reward model}
At each time step, the DT selects a control action $u_t\in\lbrace\text{DN, RO, MR, PR}\rbrace$. Choosing a DN or RO action yields a positive reward but also entails the risk of structural deterioration. Conversely, the MR and PR actions mitigate structural health deterioration but incur a negative reward. The costs associated with the structural health state, encoded in $D_t$, and the costs related to the control actions $U_t$, appearing in Equation~\eqref{eq:reward}, are defined as follows:  
\begin{equation}
    R_t^{\text{health}}(d) =
    \begin{cases}
        0 & \text{if $y = 0$}, \delta =0, \\
        -\exp(5\delta)+4 & \text{if } y\neq0, 0 <\delta < 80\%,  \\
        -250 & \text{if $\delta \geq 80\%$},
    \end{cases}\qquad
    R_t^{\text{control}}(u) = 
    \begin{cases}
        +30 & \text{if $u$ = DN}, \\
        +18 & \text{if $u$ = RO}, \\
        -75 & \text{if $u$ = MR}, \\
        -250 & \text{if $u$ = PM}.
    \end{cases}
\end{equation}
These non-dimensional rewards represent indicative costs charged to the decision-maker. Actual values can be derived from service and cost lists provided by state agencies and companies~\cite{papakonstantinou2014planning2}. For instance, $R^\text{health}$ is designed to penalize the progressive deterioration of the structural condition as a function of $\delta$. These costs may reflect various impacts, such as reduction in service level, workplace accidents, or increased structural failure probability~\cite{papakonstantinou2014planning2}, with the last entry imposing a significantly negative reward for excessively compromised structural states.

\subsection{Results}
In this section, we present the results of DT simulations carried out over an expected lifespan of $T=60$ time steps. At each time step, new observational data are simulated based on the (unknown) ground truth structural health and the most recently applied control action. The DT assimilates these data, estimates the digital state, and recommends the next control action.
\begin{figure}[!ht]
\centering
  \includegraphics[width=.95\linewidth]{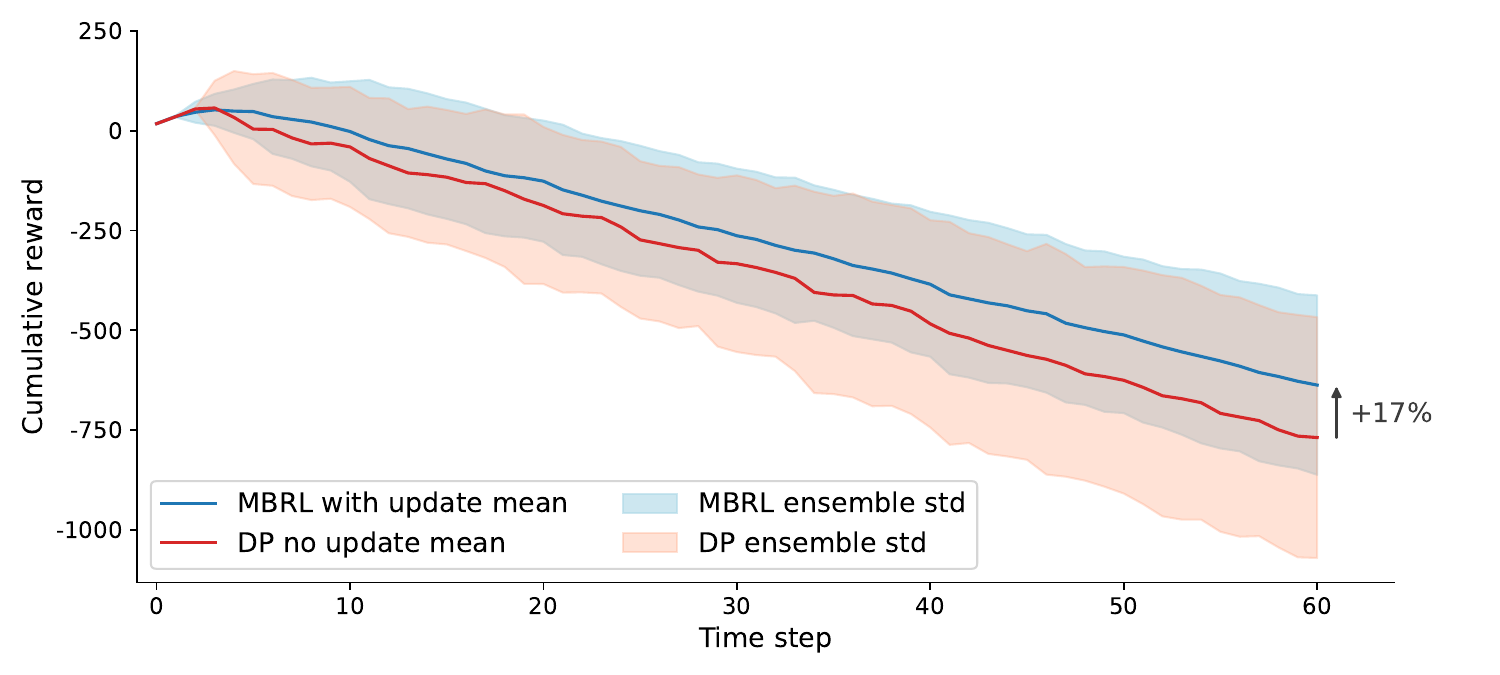}
  \caption{Digital twin online phase - Comparison of cumulative rewards under control policies computed using finite-horizon model-based reinforcement learning (MBRL) with precision updates and finite-horizon dynamic programming (DP) without precision updates. Solid lines denote the mean cumulative rewards, while the shaded areas indicate the one-standard-deviation credibility intervals. Results are shown for two clusters of simulations, each comprising $100$ runs initialized with different random seed.}
  \label{fig:MBRL_vs_DP}
\end{figure}

\begin{figure}[ht]
\centering
  \includegraphics[width=.9\linewidth]{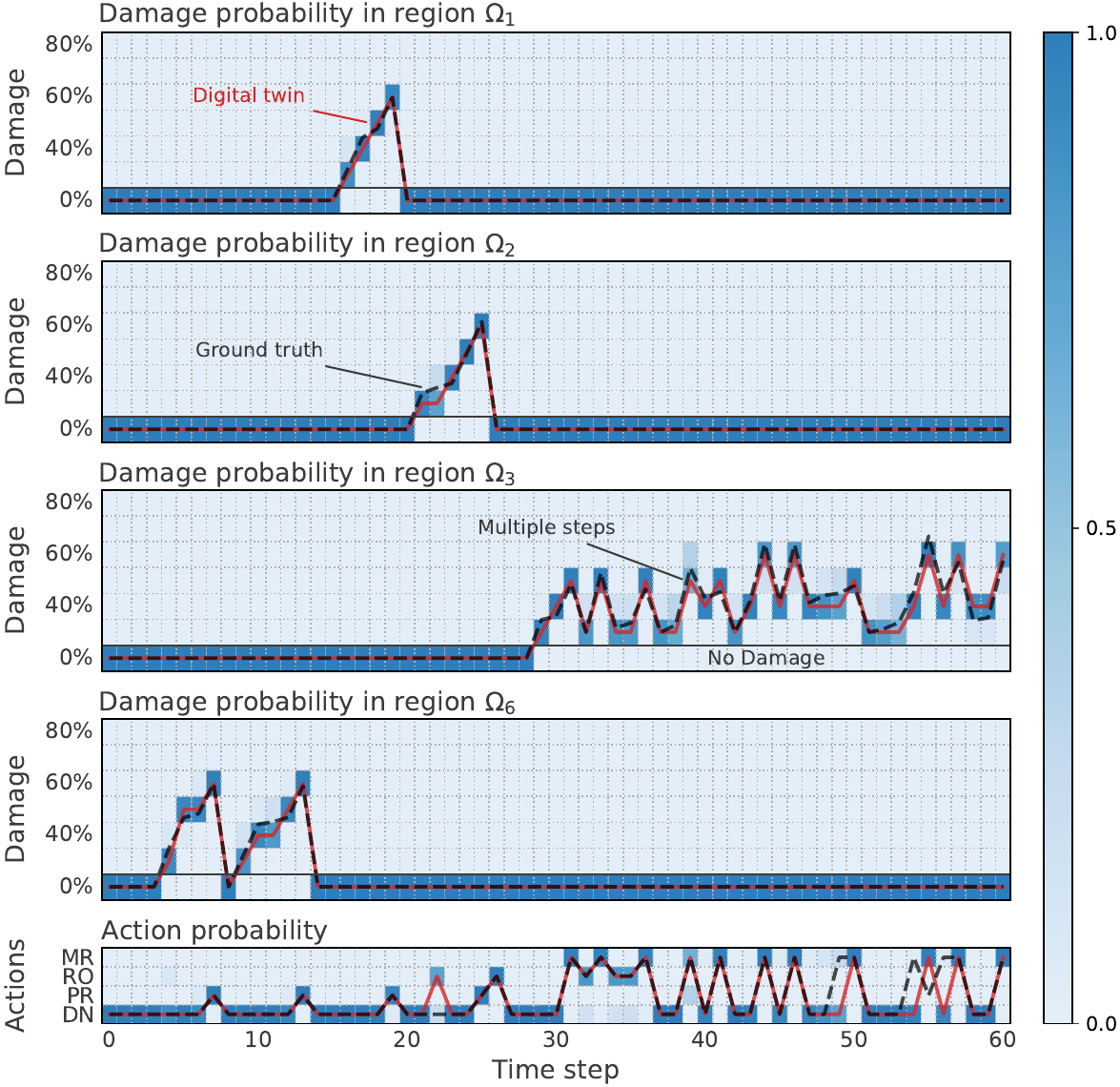}
  \caption{Digital twin online phase - Simulation under finite-horizon model-based reinforcement learning with precision updates. Probabilistic and best point estimates of: (top four panels) digital state evolution against the ground truth physical state for regions $\Omega_1$, $\Omega_2$, $\Omega_3$, and $\Omega_6$; (bottom panel) control actions informed by the digital twin against the optimal action under the ground truth. The background colors represent the digital state beliefs in the top panels, and the control action beliefs in the bottom panel.}
  \label{fig:finite_MBRL}
\end{figure}

By running a cluster of $100$ simulations, each spanning $60$ time steps and initialized with a different random seed for the ground-truth physical process, both the DT using a finite-horizon dynamic programming policy without precision updates --- computed based on the initial prior transition model --- and the adaptive DT using a finite-horizon MBRL policy with precision updates, where control policies are iteratively recomputed using the updated transition model, exhibit zero failures. Figure~\ref{fig:MBRL_vs_DP} compares the evolution of cumulative rewards, showing that the adaptive DT yields significantly higher cumulative rewards (+17\%). This improvement stems from the online refinement of the transition model, which enables more informed planning and decision-making. Moreover, our experiments indicate that non-adaptive policies can lead to undesirable outcomes; depending on the interplay between the transition model, the reward model, and the ground-truth dynamics, such policies may cause the DT to fail, i.e., the structural state reaches a critically compromised condition. For instance, digital failure (corresponding to $\delta\geq80\%$ and symbolizing structural collapse) can occur if the initial prior over the transition dynamics is poorly aligned with the ground-truth process. In contrast, the robustness of MBRL ensures that the DT consistently maintains the structural health above critical thresholds.

In-simulation performance is further assessed through a tracking accuracy metric based on the maximum a-posteriori estimate of the digital state relative to the ground-truth physical process. To account for the $\delta$ discretization, accuracy is computed using a $10\%$ tolerance, corresponding to the discretization step. Under these conditions, the adaptive DT achieves a mean accuracy of $93.6\%$ with standard deviation of $4\%$, while the non-adaptive counterpart attains a mean accuracy of $93.15\%$ with standard deviation of $4.3\%$. To avoid biasing the metric, true undamaged cases are excluded from this computation, as they are easier to identify and would favor conservative policies that promote occupancy of the undamaged state. The similar tracking accuracies observed for the two configurations are expected, since the observation model itself is not modified by the adaptive updates.

\begin{figure}[!ht]
\centering
  \includegraphics[width=.75 \linewidth]{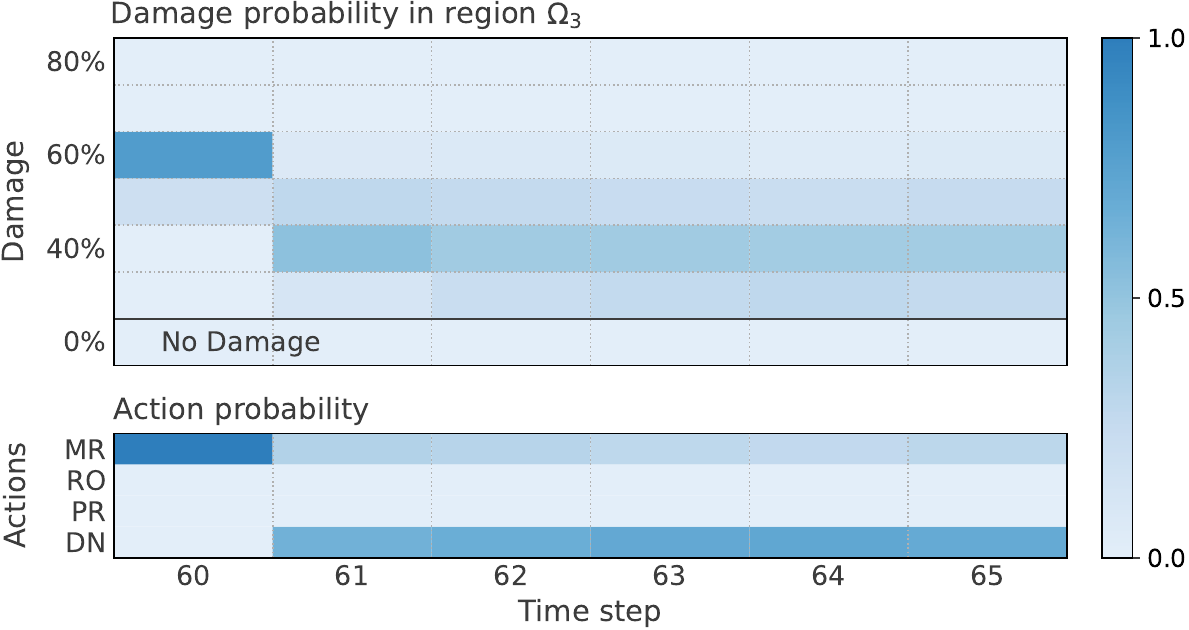}
  \caption{Digital twin online phase - Future predictions starting at $t_c=60$ under finite-horizon model-based reinforcement learning with precision updates. The background colors represent the digital state beliefs in the top panels, and the control action beliefs in the bottom panel.} 
  \label{fig:prediction}
\end{figure}

Figure~\ref{fig:finite_MBRL} illustrates a sample DT simulation using finite-horizon MBRL with precision updates. The DT accurately tracks the digital state evolution with relatively low uncertainty and a limited delay of at most two time steps with respect to the ground truth. This delay is attributed both to the need to update the prior belief based on data from previous time steps and to occasional misclassification by the DL model between digital states associated with the same damage location but adjacent $\delta$ intervals. The figure also demonstrates how accurate inference of transition dynamics and optimization of finite-horizon control policies allow the PR action to be avoided as the simulation approaches the end of the planning horizon (i.e., the decommissioning time). Specifically, the belief of selecting a PR action for the same damage magnitude progressively decreases as the simulation advances. Additionally, Figure~\ref{fig:prediction} shows the predicted evolution of the digital state and control actions starting from $t_c=60$ and spanning a prediction horizon of $t_p = t_c+5$. The DT firsts forecasts the expected improvement in structural health after taking the MR action, based on the learned transition model. This followed by a high-probability prediction of DN and MR actions, yielding a pseudo-stable belief propagation pattern that maintains most of the probability mass of the digital state within the first three $\delta$ intervals.

\begin{figure}[!ht]
\centering
  \includegraphics[trim=30 30 20 30, clip, width=0.95\linewidth]{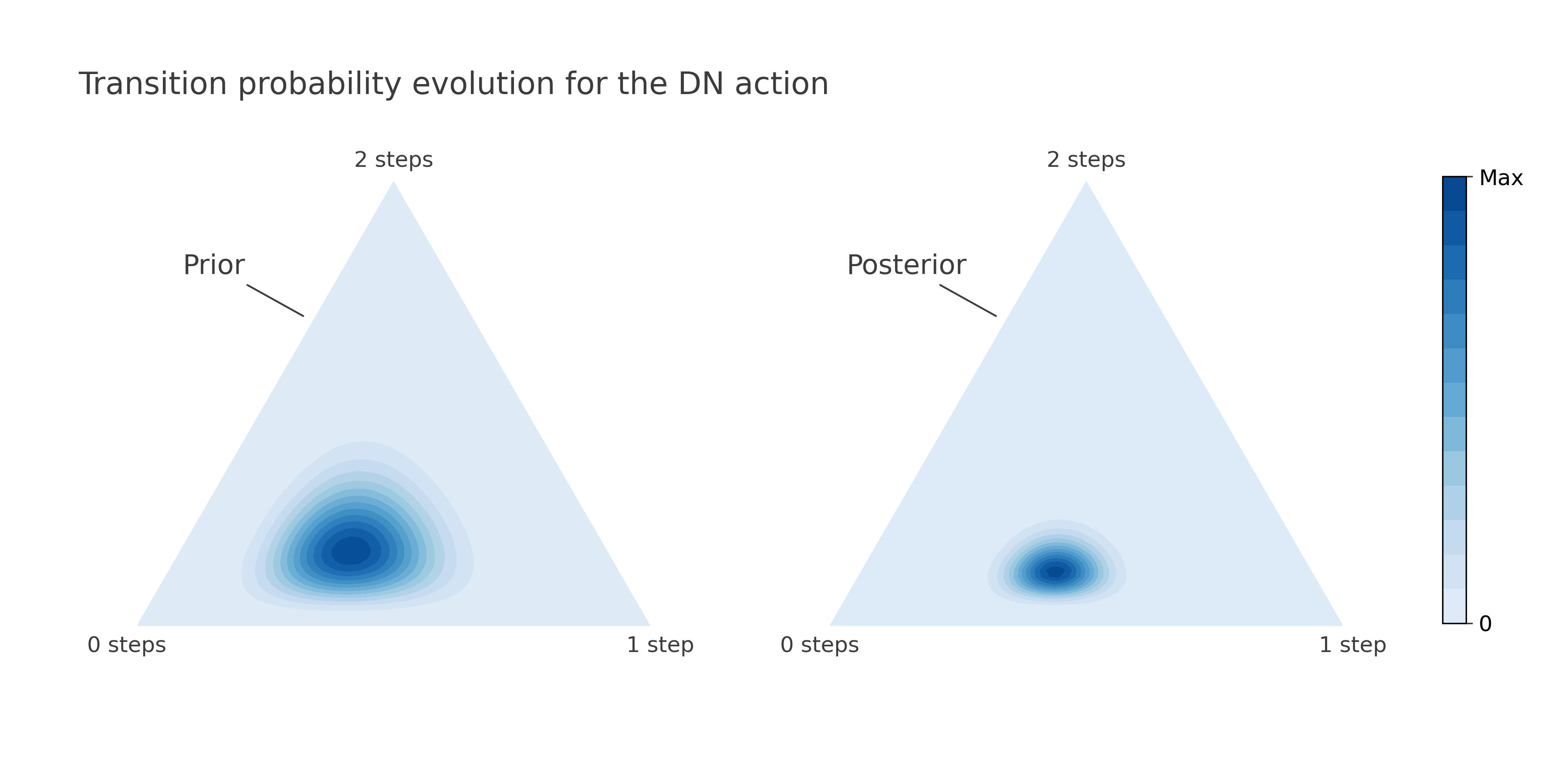}
    \\
  \includegraphics[trim=30 50 20 30, clip, width=0.95\linewidth]{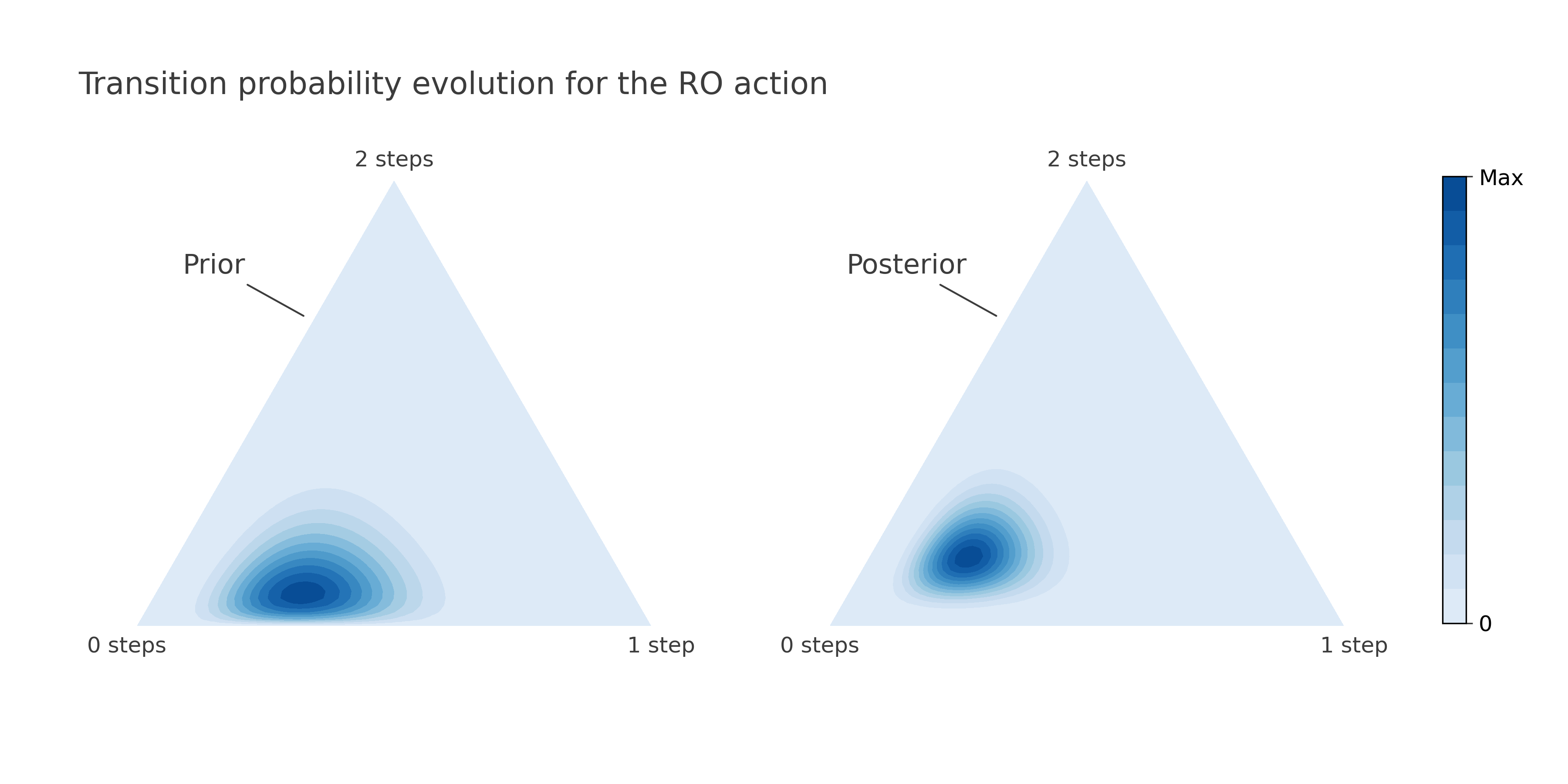}
  \\
  \includegraphics[trim=30 30 20 30, clip, width=0.95\linewidth]{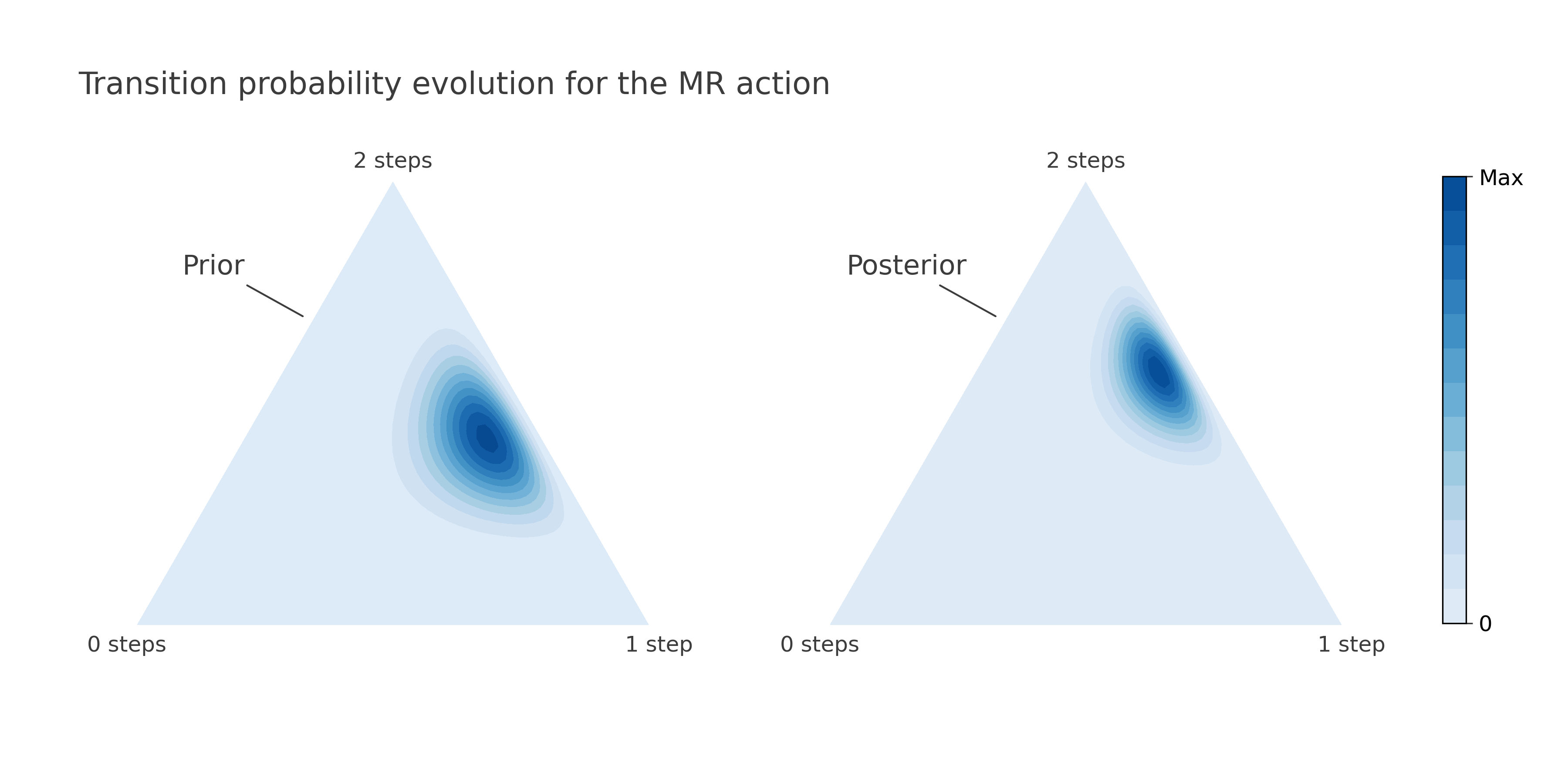}
  \caption{Prior and posterior distributions of the transition probabilities associated with the do nothing action (first row), the restrict operations action (second row), and the minor repair action (third row). Initial priors correspond to $t=0$, while the posterior distributions correspond to~$t=60$.}
  \label{fig:betas}
\end{figure}

Figure~\ref{fig:betas} illustrates the initial Dirichlet priors on the transition probabilities for the DN, RO, and MR actions, together with the corresponding posteriors at the end of the simulation. In all cases, the posteriors exhibit clear convergence toward the (unknown) ground-truth transition probabilities while significantly reducing uncertainty. The corresponding posterior parameters~read:
\begin{align}
    \overline{\boldsymbol{\alpha}}_\text{DN}=&(\alpha_\text{DN}^0+N_\text{DN}^0, \alpha_\text{DN}^1+N_\text{DN}^1, \alpha_\text{DN}^2+N_\text{DN}^2)^\top,\\
        \overline{\boldsymbol{\alpha}}_\text{RO}=&(\alpha_\text{RO}^0+N_\text{RO}^0, \alpha_\text{RO}^1+N_\text{RO}^1, \alpha_\text{RO}^2+N_\text{RO}^2)^\top,\\
            \overline{\boldsymbol{\alpha}}_\text{MR}=&(\alpha_\text{MR}^0+N_\text{MR}^0, \alpha_\text{MR}^1+N_\text{MR}^1, \alpha_\text{MR}^2+N_\text{MR}^2)^\top,
\end{align}
whose probability masses are centered around the true modes, indicating that the updated beliefs tend to reflect the actual transition probabilities while reducing uncertainty.

A dedicated sensitivity analysis to observation uncertainty is not repeated in the present work; instead, we refer the reader to~\cite{MT_AIF}, where this aspect has been systematically investigated. The analysis considers progressively degraded observation conditions, including incomplete data streams during train passages, varying observation failure rates, and uncertainty in damage extent. The results show a gradual performance degradation: reduced data availability increases state uncertainty and may delay optimal actions, while the transition model helps preserve acceptable performance even under severe data loss, although with increased uncertainty. On the other hand, uncertainty in damage extent mainly affects estimation accuracy, with negligible impact on failure rates.

\section{Conclusion}
\label{sec:conclusion}
In this paper, we have introduced adaptivity into the predictive digital twin framework for structural health monitoring proposed in~\cite{Torzoni_DT} to enhance decision-making capabilities. The framework models the bi-directional information flow between the monitored asset and its virtual replica using dynamic Bayesian networks. Adaptivity is achieved by treating the state transition probabilities as random variables endowed with conjugate prior distributions that are updated online through Bayesian inference. Adaptive planning is integrated via parametric Markov decision processes, enabling dynamic control policies with precision updates. This work extends the results in~\cite{tezzele2024adaptive} to a broader class of distributions, thereby allowing the proposed DT framework to be applied to real-life, large-scale systems.

The computational procedure has been assessed through simulations of monitoring, management, and maintenance planning for the H{\"o}rnefors railway bridge. The DT effectively tracks the structural health evolution, recommending timely and appropriate control actions with low uncertainty. By continuously refining the state transition models using online data, the DT dynamically adapts to evolving structural conditions, providing more reliable structural health assessments, improved forecasts accuracy, and enhanced decision-making. When discrepancies with the ground truth arise, the DT demonstrates the ability to self-correct after assimilating new observational~data.

Future research will address additional challenges, such as environmental uncertainties that can cause sudden structural degradation~\cite{sanchez2011life,guo2020two}, complementing the long-term deterioration effects considered in this study. Particular attention will be devoted to extreme events driven by environmental risks and climate change, which require robust control policies that proactively account for the probability of such events. Further efforts will focus on partial observability by incorporating information-gathering actions, such as installing new sensors and scheduling inspections. In addition, future work will explore explicitly non-stationary dynamics, such as time-dependent deterioration rates or abrupt changes in system behavior. Addressing these scenarios may require the integration of adaptive learning strategies, such as forgetting mechanisms or context-aware models~\cite{gama2014survey,padakandla2021survey}, to effectively track evolving transition dynamics.

\section*{Acknowledgements}
M.To. acknowledges the financial support from the Politecnico di Milano through the interdisciplinary Ph.D. Grant ``Physics-informed deep learning for structural health monitoring''. M.To. and A.M. acknowledge the PRIN 2022 Project UQ-FLY (Numerical approximation of uncertainty quantification problems for PDEs by multi-fidelity methods), funded by the NextGenerationEU program within the PNRR scheme (M4C2, Investment 1.1). A.M. acknowledges the Project “Reduced Order Modeling and Deep Learning for the real-time approximation of PDEs (DREAM)” (Starting Grant No. FIS00003154), funded by the Italian Science Fund (FIS) - Ministero dell’Università e della Ricerca and the project FAIR (Future Artificial Intelligence Research), funded by the NextGenerationEU program within the PNRR-PE-AI scheme (M4C2, Investment 1.3, Line on Artificial Intelligence). M.To. and A.M. acknowledge the financial support from the ERC advanced grant IMMENSE (Grant Agreement 101140720), funded by the European Union. Views and opinions expressed are however those of the authors only and do not necessarily reflect those of the European Union or the European Research Council Executive Agency. Neither the European Union nor the granting authority can be held responsible for them.

\section*{Data Availability}
The DT framework was implemented using the \texttt{pgmtwin} Python package \cite{pgmtwin}. The observational data used to run the experiments presented in Section~\ref{sec:res} are available in the public repository \texttt{digital-twin-SHM}~\cite{repo}. The DL models trained according to the implementation details reported in~\cite{Torzoni_DT} are also made available in the same repository. The \texttt{Matlab} library for finite element simulation and reduced-order modeling of partial differential equations used to generate these data is available in the public repository \texttt{Redbkit}~\cite{Redbkit}.

\bibliographystyle{abbrvurl}
\bibliography{dt-transitions-biblio.bib}

\appendix

\section{State-dependent Dirichlet-Multinomial parametrization}
\label{sec:state_dep}
In this Appendix, we provide the formulation and update scheme for the general case of a state-dependent Dirichlet-Multinomial transition model. In such a case, $\boldsymbol{\theta}_{d,u}=(\theta_{d,u}^{1},\ldots,\theta_{d,u}^{|\mathcal{D}|-1})$ is defined over the $|\mathcal{D}|-1$ probability simplex. The Dirichlet prior over it is given by:
\begin{equation}
\label{eq:prior}
p(\boldsymbol{\theta}_{d,u})=\frac{1}{B(\boldsymbol{\alpha}_{d,u})}\prod_{k=1}^{|\mathcal{D}|}(\theta_{d,u}^{k})^{\alpha_{d,u}^{k} - 1},
\end{equation}   
where $B(\boldsymbol{\alpha}_{d,u})$ is the multivariate Beta function that ensures normalization:
\begin{equation}
B(\boldsymbol{\alpha}_{d,u})=\frac{\prod_{k=1}^{|\mathcal{D}|} \Gamma(\alpha_{d,u}^{k})}{\Gamma(\sum_{k=1}^{|\mathcal{D}|} \alpha_{d,u}^{k})},
\end{equation} 
with $\Gamma(z > 0) = \int_0^\infty a^{z-1} e^{-a} \, da$ being the Gamma function. 

Let $\mathbf{y}_{d,u}=(N^{1}_{d,u},\ldots,N^{|\mathcal{D}|}_{d,u})^\top\in\mathbb{N}^{|\mathcal{D}|}$ denote the counts of transitions from state $d$ to each state $d'$ under action $u$, collected from $\lbrace d_{0:t_c},u_{0:t_c-1}\rbrace$. Each entry is given as $N^{d'}_{d,u}=\sum_{t=0}^{t_c-1}\mathds{1}_{\lbrace d_t=d, u_t=u, d_{t+1}=d'\rbrace}$, while $N_{d,u}=\sum_{k=1}^{|\mathcal{D}|} N^{k}_{d,u}$ is the total number of transitions from state $d$ under action $u$.

The transition counts $\mathbf{y}_{d,u}$ conditioned on $\boldsymbol{\theta}_{d,u}$ follow a Multinomial distribution:
\begin{align}
\mathbf{y}_{d,u} \mid \boldsymbol{\theta}_{d,u} &\sim \text{Multinomial}(N_{d,u}, \boldsymbol{\theta}_{d,u}),\\
p(\mathbf{y}_{d,u}\mid  \boldsymbol{\theta}_{d,u}) &\propto \prod_{k=1}^{|\mathcal{D}|} (\theta_{d,u}^{k})^{N^{k}_{d,u}}.\label{eq:likelihood}
\end{align}

By applying Bayes' theorem and substituting the likelihood~\eqref{eq:likelihood} and the prior~\eqref{eq:prior}, the unnormalized posterior distribution over $\boldsymbol{\theta}_{d,u}$ is given by:
\begin{equation}
p(\boldsymbol{\theta}_{d,u} \mid \mathbf{y}_{d,u}) \propto p(\mathbf{y}_{d,u}\mid  \boldsymbol{\theta}_{d,u})p(\boldsymbol{\theta}_{d,u}) = 
\prod_{k=1}^{|\mathcal{D}|} (\theta_{d,u}^{k})^{N^{k}_{d,u}}\cdot\prod_{k=1}^{|\mathcal{D}|} (\theta_{d,u}^{k})^{\alpha_{d,u}^{k} - 1} = \prod_{k=1}^{|\mathcal{D}|} (\theta_{d,u}^{k})^{\alpha_{d,u}^{k} - 1 + N^{k}_{d,u}},
\end{equation}
which remains Dirichlet distributed:
\begin{equation}
p(\boldsymbol{\theta}_{d,u} \mid \mathbf{y}_{d,u}) = \text{Dirichlet}(\overline{\boldsymbol{\alpha}}_{d,u}),
\end{equation}
with updated concentration parameters $\overline{\boldsymbol{\alpha}}_{d,u} = (\overline{\alpha}_{d,u}^{1}, \ldots, \overline{\alpha}_{d,u}^{|\mathcal{D}|})^\top\in\mathbb{R}^{|\mathcal{D}|}$, computed as:
\begin{equation}
\overline{\alpha}_{d,u}^{k} = \alpha_{d,u}^{k} + N^{k}_{d,u}, \quad k=1,\ldots,|\mathcal{D}|.
\end{equation}

\section{State-independent Beta-Bernoulli parametrization}
\label{sec:beta_bern}
In this Appendix, we provide the formulation and update scheme for a state-independent Beta-Bernoulli transition model, parameterized by a 1-step transition probability $p_u\in\mathbb{R}$. This model constitutes a special case of the Dirichlet-Multinomial framework for state-independent transitions. When applied to the illustrative example introduced in Section~\ref{subsec:online_learning}, it yields the following transition matrix:
\begin{equation}
\mathcal{P}_u(p_u) = 
\begin{pmatrix}
1-p_u & 0 & 0 & 0 & 0\\
p_u & 1-p_u & 0 & 0 & 0\\
0 & p_u & 1-p_u & 0 & 0\\
0 & 0 & p_u & 1-p_u & 0\\
0 & 0 & 0 & p_u & 1\\
\end{pmatrix},
\end{equation}
where the system transitions from state $d$ to the subsequent state $d'$ with probability $p_u$, and remains in the same state with probability $1-p_u$. Analogously to the Dirichlet-Multinomial case, the online learning procedure for the Beta-Binomial transition model can be formalized as follows:
    \begin{align}
        p_u &\sim \text{Beta}(a_u, b_u), \\
        \mathbf{y}_u \mid  p_u &\sim \text{Binomial}(N_u,p_u).
    \end{align}
Herein, $p_u$ is assigned a Beta prior with shape hyperparameters $a_u,b_u\in\mathbb{R}$ for action $u$. The vector $\mathbf{y}_u=(N^{0}_{u},N^{1}_{u})^\top$ collects the observed transitions counts $N^{0}_{u} = \sum_{t=0}^{t_c-1}\mathds{1}_{\lbrace d_t = d_{t+1}, u_{t} = u\rbrace}$ and \mbox{$N^{1}_{u} = \sum_{t=0}^{t_c-1}\mathds{1}_{\lbrace d_t \neq d_{t+1}, u_{t} = u\rbrace}$}; the likelihood term is modeled as a binomial distribution with \mbox{$N_u=\sum_{t=0}^{t_c-1}\mathds{1}_{\lbrace u_{t} = u\rbrace}$}. The posterior distribution of $p_u$ is then updated as:
\begin{equation}
        p_u \mid  \mathbf{y}_u\sim \text{Beta}(a_u + N^{1}_{u}, b_u + N^{0}_{u}).
\end{equation}
While this transition model does not account for multi-step transitions --- which can be restrictive for systems with multiple components deteriorating at varying rates --- a state-independent 1-step transition model can be highly efficient in scenarios characterized by slow deterioration rates. 

\end{document}